\documentclass[sigconf]{acmart}

\usepackage{subcaption}
\usepackage{multirow}
\usepackage{cleveref}
\usepackage{enumitem}
\usepackage{arydshln}

\AtBeginDocument{%
  }

\copyrightyear{2025}
\acmYear{2025}
\setcopyright{acmlicensed}
\acmConference[MM '25] {Proceedings of the 33rd ACM International Conference on Multimedia}{October 27--31, 2025}{Dublin, Ireland.}
\acmBooktitle{Proceedings of the 33rd ACM International Conference on Multimedia (MM '25), October 27--31, 2025, Dublin, Ireland}
\acmISBN{979-8-4007-2035-2/2025/10}
\acmDOI{10.1145/3746027.3755195}

\begin{document}

\title{Multiple Queries with Multiple Keys: A Precise Prompt Matching Paradigm for Prompt-based Continual Learning}


\author{Dunwei Tu}
\email{tudunwei@smail.nju.edu.cn}
\affiliation{%
  \institution{National Key Laboratory for Novel Software Technology, Nanjing University}
  \city{Nanjing}
  \country{China}}
  
\author{Huiyu Yi}
\email{211300010@smail.nju.edu.cn}
\affiliation{%
  \institution{National Key Laboratory for Novel Software Technology, Nanjing University}
  \city{Nanjing}
  \country{China}}

\author{Yuchi Wang}
\email{yuchi.wang@smail.nju.edu.cn}
\affiliation{%
  \institution{National Key Laboratory for Novel Software Technology, Nanjing University}
  \city{Nanjing}
  \country{China}}

\author{Baile Xu}
\authornote{Baile Xu is the corresponding author.}
\email{xubaile@nju.edu.cn}
\affiliation{%
  \institution{National Key Laboratory for Novel Software Technology, Nanjing University}
  \city{Nanjing}
  \country{China}}

\author{Jian Zhao}
\email{jianzhao@nju.edu.cn}
\affiliation{%
  \institution{School of Electronic Science and Engineering, Nanjing University}
  \city{Nanjing}
  \country{China}}

\author{Furao Shen}
\email{frshen@nju.edu.cn}
\affiliation{%
  \institution{National Key Laboratory for Novel Software Technology, Nanjing University}
  \city{Nanjing}
  \country{China}}







\renewcommand{\shortauthors}{Denwei Tu et al.}

\begin{abstract}
Continual learning requires machine learning models to continuously acquire new knowledge in dynamic environments while avoiding the forgetting of previous knowledge. Prompt-based continual learning methods effectively address the issue of catastrophic forgetting through prompt expansion and selection. However, existing approaches often suffer from low accuracy in prompt selection, which can result in the model receiving biased knowledge and making biased predictions. To address this issue, we propose the Multiple Queries with Multiple Keys (MQMK) prompt matching paradigm for precise prompt selection. The goal of MQMK is to select the prompts whose training data distribution most closely matches that of the test sample. Specifically, Multiple Queries enable precise breadth search by introducing task-specific knowledge, while Multiple Keys perform deep search by representing the feature distribution of training samples at a fine-grained level. Each query is designed to perform local matching with a designated task to reduce interference across queries. Experiments show that MQMK enhances the prompt matching rate by over 30\% in challenging scenarios and achieves state-of-the-art performance on three widely adopted continual learning benchmarks. The code is available at \url{https://github.com/DunweiTu/MQMK}.

\end{abstract}




\maketitle

\begin{figure}[t]
  \centering
  \begin{subfigure}{\linewidth}
  {\includegraphics[width=\textwidth,trim=160 120 160 120,clip]{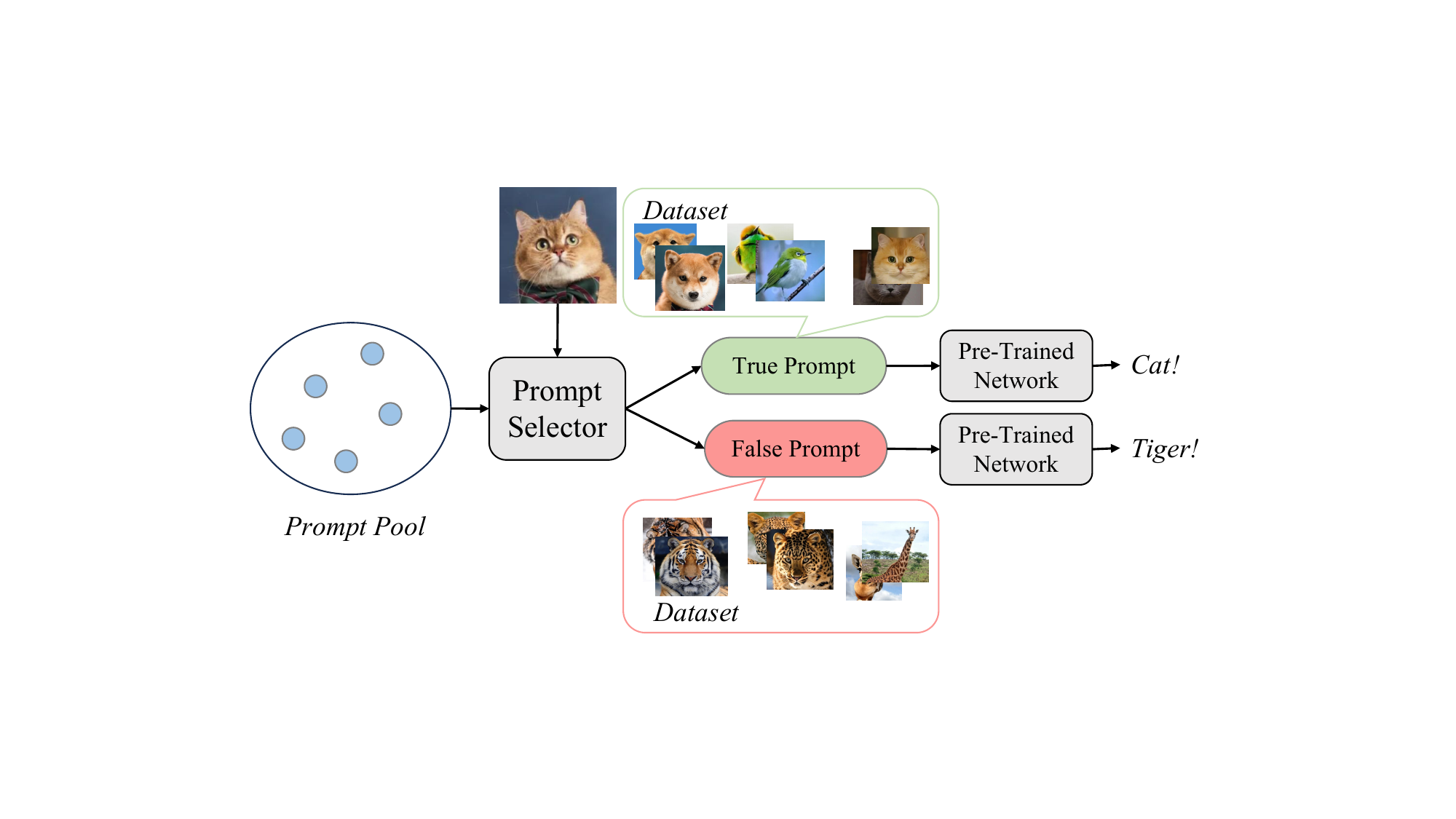}}
  \caption{The motivation for selecting the appropriate prompt.}
  \label{fig:motivation1}
  \end{subfigure}
  \begin{subfigure}{\linewidth}
  {\includegraphics[width=\textwidth,trim=80 140 80 140,clip]{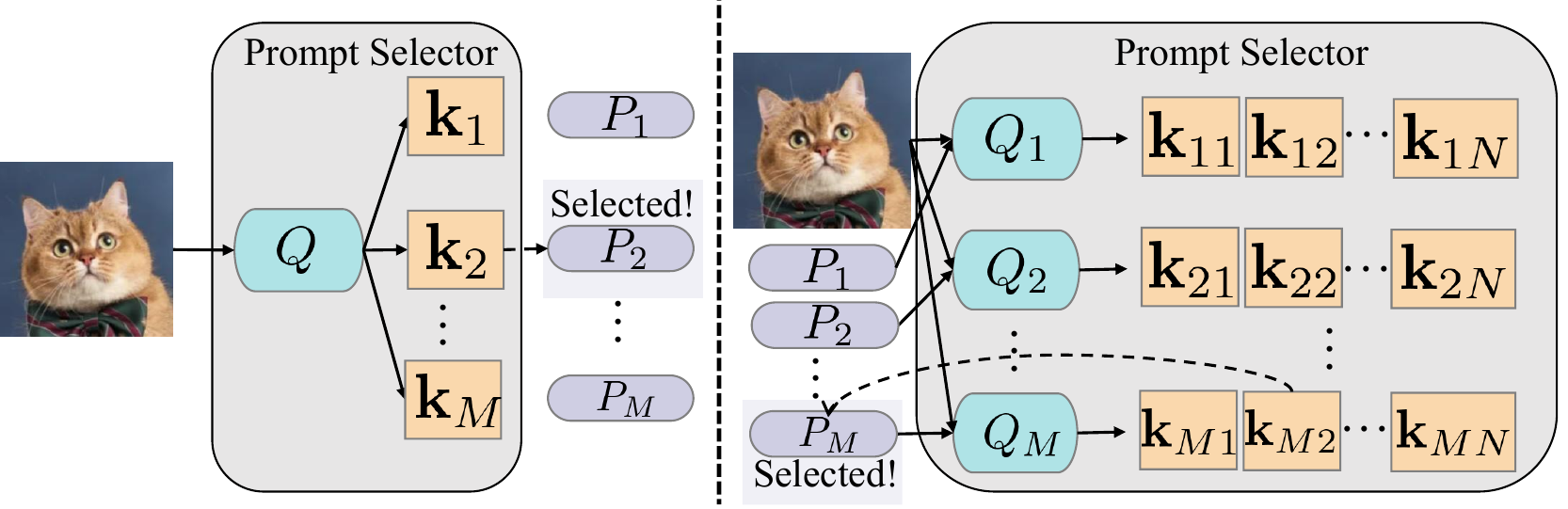}}
  \caption{\textbf{Left:} Prior work. \textbf{Right:} Our work.}
  \label{fig:motivation2}
  \end{subfigure}
\caption{(a) Prompts inconsistent with test samples may introduce bias. (b) Prior prompt selector uses a single query and task-level keys; our method adopts task-level breadth queries and class-level depth keys, involving prompts in the selection process.}
  \label{fig:motivation}
\end{figure}

\section{Introduction}

Neural networks have become one of the most important models in the field of machine learning \cite{rumelhart1986learning,lecun1998gradient,graves2012long}. In real-world scenarios, data may not always be fully accessible from the start, and the dynamic nature of the world continuously introduces new data. This requires neural networks to possess the ability for continual learning \cite{belouadah2021comprehensive,zhou2024continual,de2021continual} to maintain knowledge updates and adapt to new environments. Additionally, humans do not forget old knowledge when learning new knowledge, making continual learning a key technology for building artificial intelligence that mirrors human learning.

In the context of continual learning, the data across different tasks is often non-independent and identically distributed (non-i.i.d.). Non-i.i.d. data requires the model to continuously adjust its parameters based on new data to establish decision boundaries more suited to the current data. However, this adaptation can lead to forgetting of previously learned knowledge. Some studies suggest that unless certain measures are taken, the neural network's ability to recognize previously encountered classes inevitably deteriorates during learning of new tasks, a phenomenon known as catastrophic forgetting \cite{nguyen2019toward,mccloskey1989catastrophic,mcclelland1995there}.

Early continual learning methods design sustainable learning from aspects such as parameters \cite{kirkpatrick2017overcoming,li2017learning}, data \cite{rebuffi2017icarl,shin2017continual}, gradients \cite{lopez2017gradient,mirzadeh2020understanding}, features \cite{madaan2021representational,pham2021dualnet}, and architectures \cite{yan2021dynamically,wang2022foster}, training a neural network from scratch, which has yielded promising results. Recent continual learning algorithms \cite{zhou2024continual,wang2022learning,mcdonnell2024ranpac,tu2025embedding,yi2024few} have started to focus on using pre-trained models to leverage their strong generalization ability for adapting to dynamic downstream tasks.

Prompt-based methods \cite{wang2022dualprompt,wang2022learning,smith2023coda,jung2023generating,gao2024consistent,kurniawan2024evolving,yang2024generating,feng2024cp}, as emerging pre-trained model–based approaches, counter forgetting by cleverly reusing the generalization knowledge of pre-trained models. These methods utilize a frozen backbone network, such as Vision Transformer (ViT) \cite{dosovitskiy2020image} pre-trained on large-scale datasets like ImageNet \cite{russakovsky2015imagenet}, and adapt to continual learning tasks through Visual Prompt Tuning (VPT) \cite{jia2022visual} or Prefix Tuning \cite{li2021prefix}. A major advantage is that task-specific knowledge is stored in prompts, and as tasks increase, prompts are expanded to form a prompt pool covering all learned knowledge.

Existing methods \cite{wang2022learning,wang2022dualprompt} use a query-key matching mechanism for prompt selection: a pre-trained ViT without prompts as the query function, and task-specific learnable parameters as the key. However, the matching is often imprecise. A prompt trained on a distribution inconsistent with test samples may fail to assist inference (\cref{fig:motivation1}). Results show that improving the matching rate effectively boosts model performance (see \cref{Matching rate}).

To improve the matching rate and consistency between prompt and test samples, we propose the Multiple Queries with Multiple Keys (MQMK) paradigm. Unlike the current Single Query–Single Key (SQSK) paradigm \cite{wang2022learning,wang2022dualprompt,smith2023coda}, our method builds a query pool—each query corresponding to a task—for task-level breadth search (\cref{fig:motivation2}). Additionally, we extend task-level keys to Multiple Keys within each task for deep search. Experiments show performance is best when keys are at class level. Due to knowledge discrepancies among queries, we introduce a local matching mechanism to mitigate mutual interference.
To address query overhead under resource constraints, we propose MQMK-Efficient Inference (MQMK-EI), which uses a single enhanced query to match multiple keys. By horizontally expanding the query-key matching mechanism, MQMK achieves up to 32.82\% improvement in matching rate and delivers SOTA performance across three recognized continual learning datasets. Our results show that, beyond improving prompt quality, directly enhancing the matching mechanism can effectively improve performance.

The contributions of this paper are summarized as follows:
\setlist[enumerate]{leftmargin=1.2em, labelwidth=0em}
\begin{enumerate}
\item We highlight the importance of query-key matching rate in prompt-based methods, define current approaches as SQSK, and introduce the MQMK paradigm, which achieves 32.82\% improvement and SOTA performance.
\item We identify that SQSK lacks task-specific knowledge in the query, leading to low precision, and that the prompt itself is not involved in prompt selection. To address this, we extend Single Query to Multiple Queries for task-level breadth search.
\item After introducing Multiple Queries, class-level feature gaps become more pronounced.
A single key cannot fully represent training distributions. Therefore, we extend Single Key to Multiple Keys for deep search and find class-level keys perform best. Finally, we propose a local matching mechanism to further enhance matching precision.
\end{enumerate}

\section{Related Work}

\textbf{Continual Learning.}
Early continual learning methods can generally be classified into five categories: regularization-based methods \cite{kirkpatrick2017overcoming,li2017learning}, replay-based methods \cite{rebuffi2017icarl,shin2017continual}, optimization-based methods \cite{lopez2017gradient,mirzadeh2020understanding}, representation-based methods \cite{madaan2021representational,pham2021dualnet}, and architecture-based methods \cite{yan2021dynamically,wang2022foster}.
Regularization-based methods address the forgetting problem by adding explicit regularization terms to balance the old and new tasks.
Replay-based methods design strategies to retain important old samples in order to preserve the model's previous knowledge.
Optimization-based methods aim to make the gradients of new and old tasks as independent as possible, avoiding interference in order to prevent forgetting.
Representation-based methods enhance the compatibility with new knowledge through meta-learning \cite{finn2017model,ravi2017optimization} and self-supervised learning \cite{chen2020simple}.
Architecture-based methods adapt to dynamic task objectives through dynamic networks.


\textbf{Prompt-based Continual Learning.}
Prompt-based methods are a type of continuous learning method based on pre-trained models.
They leverage task-specific knowledge from the prompt and the generalized knowledge from the pre-trained model, resulting in superior performance.
L2P \cite{wang2022learning} and DualPrompt \cite{wang2022dualprompt} introduce Visual Prompt Tuning (VPT) into continual learning, where prompt vectors help mitigate catastrophic forgetting by providing task-specific conditioning while keeping the backbone frozen.
CODA-P \cite{smith2023coda} introduces a decomposed attention-based prompt querying method, where prompts are combined with varying weights.
ESN \cite{wang2023isolation} addresses stage interference and performance imbalance by using stage-isolated classifiers, energy normalization, and voting-based inference.
EvoPrompt \cite{kurniawan2024evolving} addresses prompt selection mismatches and adaptive prompting challenges by using a dynamic, evolving prompt memory system that integrates reference and working prompts through optimal transport and bias adjustment.
HiDe-Prompt \cite{wang2023hierarchical} optimizes intra-task prediction, task identity inference, and task-adaptive prediction, using task-specific prompts and contrastive regularization to overcome sub-optimality in self-supervised pretraining.
CPrompt \cite{gao2024consistent} employs a random prompt selection training approach to address the issue of prompt inconsistency between training and testing phases.
\begin{figure}
    \centering
    \includegraphics[width=0.3\textwidth,trim=0 10 0 10,clip]{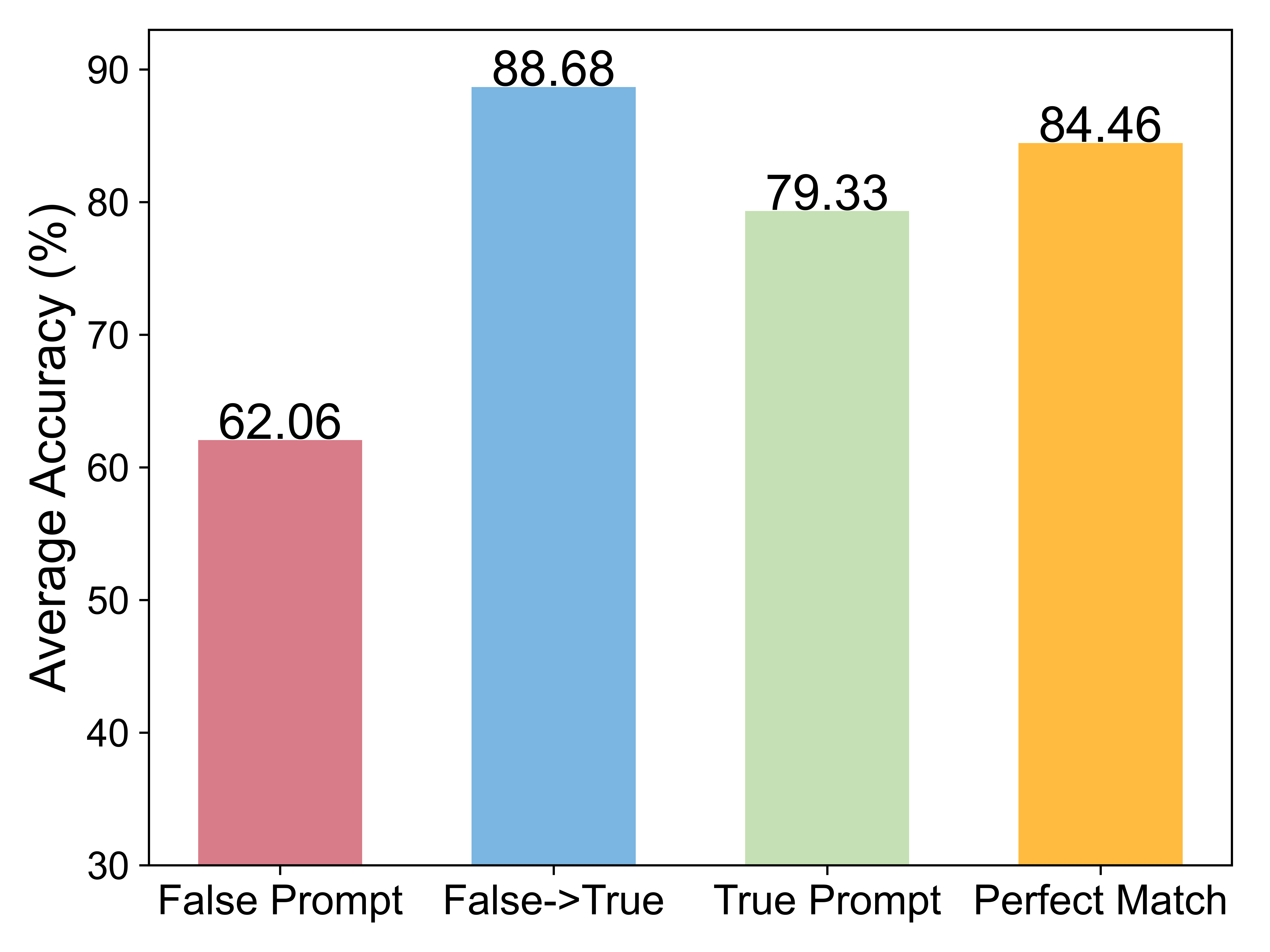}
    \caption{The average accuracy for four scenarios: when SQSK selects False Prompt and True Prompt, when samples initially with False Prompt are manually replaced with True Prompt, and when all samples use True Prompt (Perfect Match).
    }
    \label{fig:acc_true_false_perfect}
\end{figure}
\section{Preliminaries}

\subsection{Problem Setting}
In our problem setting, we feed the model with a sequence of dataset $\left\{ \mathcal{D}_{t}\right\}_{t=1}^{T}$, where $\mathcal{D}_{t}$ is the dataset of the task $t$ and $T$ is the number of all tasks.
$\mathcal{D}_{t}=\left\{ (\mathbf{x}_i,y_i)\right\}$ contains pairs of the sample $\mathbf{x}_i$ and its corresponding label $y_i$.
Each task has a distinct label space, meaning that \( \mathcal{Y}^{t} \cap \mathcal{Y}^{t^{\prime}} = \varnothing \) for any \( t \neq t^{\prime} \), where \( \mathcal{Y}^{t} \) represents the label space of task \( t \).
Moreover, once the model transitions to the next task, it cannot access any of the previous datasets.
When there are test samples for inference, the model needs to make predictions based on the label spaces of all the tasks it has encountered.
We do not provide the task ID of the sample to the model, making this a more challenging class-incremental continual learning \cite{chaudhry2018efficient} setup.
\begin{figure*}[t]
    \centering
    \includegraphics[width=0.9\textwidth,trim=40 20 60 45,clip]{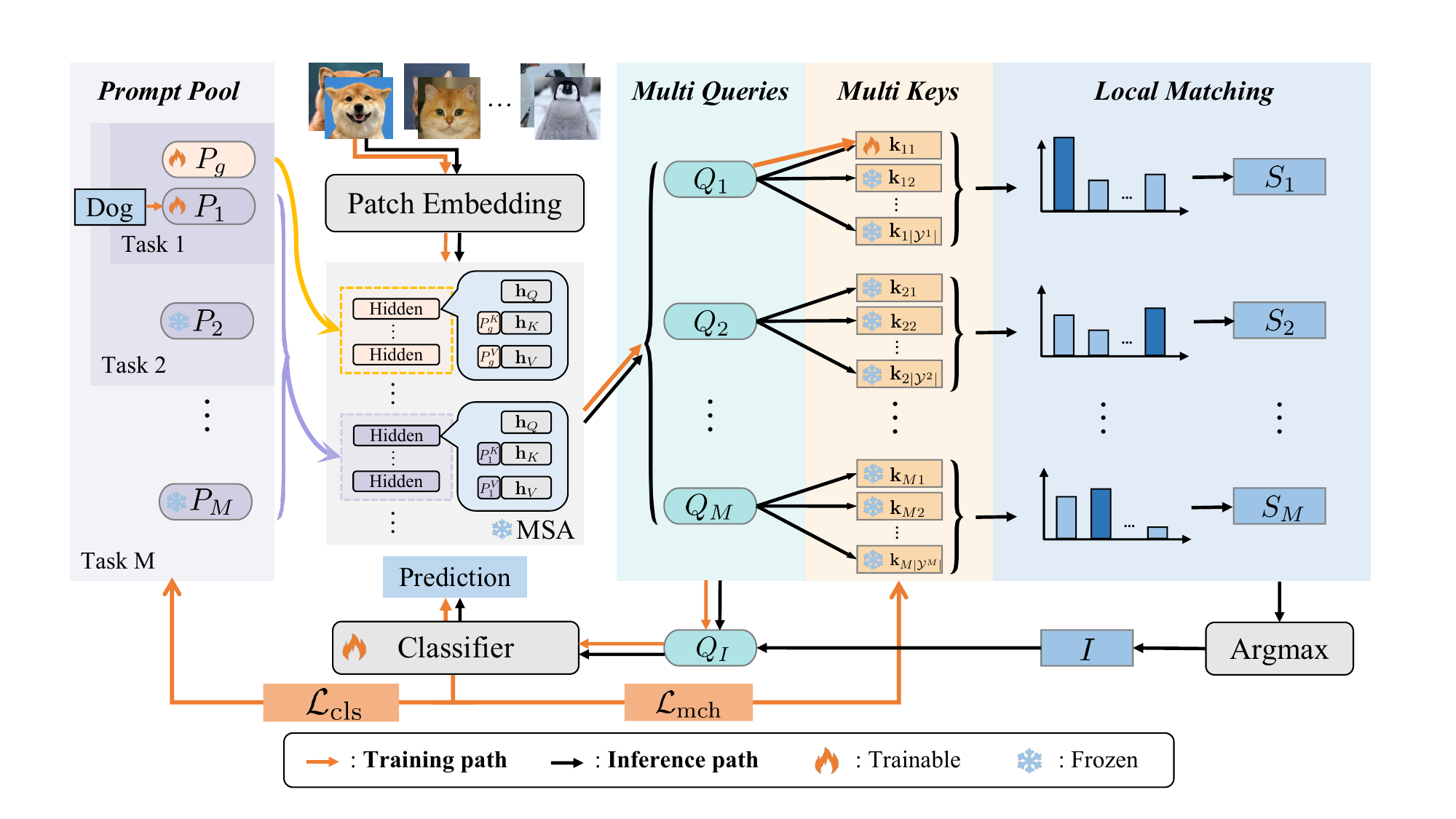}
    \caption{Overall pipeline of MQMK. \textcolor{red}{Training:} Select the true prompt based on the label information and input it into hidden states to obtain the corresponding query for classification. This process updates the prompt, the locally matched key and the classifier. \textcolor{red}{Inference:} Feed all prompts into hidden states to generate the query pool and select the most consistent query for classification by locally matching the task-level queries with the corresponding class-level keys.}
    \label{fig:main_picture}
\end{figure*}
\subsection{Model Tuning and Prompt Pool}
Following previous prompt-based continual learning works \cite{wang2022learning,wang2022dualprompt,smith2023coda,gao2024consistent,kurniawan2024evolving}, we use ViT as the backbone $f=f_r\circ f_e$ and a single linear layer $W$ as the classification head, where $f_e$ is the embedding layer and $f_r$ is the Multiple Self-Attention (MSA) layers.
The MSA layers can be written as:
\begin{align}
&\operatorname{MSA}\left(\mathbf{h}_{Q}, \mathbf{h}_{K}, \mathbf{h}_{V}\right)=\operatorname{Concat}\left(\mathrm{h}_{1}, \ldots, \mathrm{h}_{m}\right) W^{O}  \\
&\text{where } \mathrm{h}_{i}=\operatorname{Attention}\left(\mathbf{h}_{Q} W_{i}^{Q}, \mathbf{h}_{K} W_{i}^{K}, \mathbf{h}_{V} W_{i}^{V}\right),
\end{align}
where $\mathbf{h}_{Q}$, $\mathbf{h}_{K}$, and $\mathbf{h}_{V}$ are the input query, key, and value, respectively. $W^{O}$, $W_{i}^{Q}$, $W_{i}^{K}$, and $W_{i}^{V}$ are projection matrices. $m$ is the number of heads, and $i$ is the index of the layer. The function $\operatorname{Attention}(\cdot,\cdot,\cdot)$ denotes the self-attention mechanism.
Given an image input \( \mathbf{x} \), we first divide it into patches $\mathbf{x}_p \in \mathbb{R}^{L \times (S^2\times C)}$, where $L$ is the token length, $S$ is patch size and $C$ is the number of channels. 
And then pass them through $f_e$ to obtain corresponding embeddings $\mathbf{x}_e \in \mathbb{R}^{L\times D}$, where $D$ is embedding (token) dimension (768).
Next, the embeddings \( \mathbf{x}_e\) and the prompt \( P \) are fed into \( f_r \) for feature extraction to obtain the output \( f_r(P;\mathbf{x}_e) \in \mathbb{R}^{L\times D} \).
Specifically, $P$ is split into $P^K, P^V \in \mathbb{R}^{L_{P}/2\times D}$, which are prepended to $\mathbf{h}_K$ and $\mathbf{h}_V$, respectively which can written by $
    \operatorname{MSA}\left(\mathbf{h}_{Q}, [P^K;\mathbf{h}_{K}], [P^V;\mathbf{h}_{V}]\right).$
For simplicity, the layer indices of $P$, $\mathbf{h}_K$, and $\mathbf{h}_V$ are omitted.
The first token of the output, \( f_r(P;\mathbf{x}_e)[0] \in \mathbb{R}^{D} \), which serves as the [class] token, is passed through the classification head \( W \in \mathbb{R}^{D \times |\mathcal{Y}^1 \cup \mathcal{Y}^2 \cup \cdots \cup \mathcal{Y}^T|} \) to obtain the prediction, where $|\mathcal{Y}^1 \cup \mathcal{Y}^2 \cup \cdots \cup \mathcal{Y}^T|$ represents the total number of classes across all tasks.

To address catastrophic forgetting, L2P and DualPrompt design a paradigm where a prompt pool is used to adapt to multiple tasks.
Specifically, as tasks expand, the expert prompts (e-prompts) specific to each task and the general prompts (g-prompts) \cite{wang2022dualprompt} shared across tasks together form the prompt pool ${\mathbf{P}=\{P_g,P_1,P_2,\cdots,P_M \}}$, where $P_g \in \mathbb{R}^{L_g \times D}$ is the g-prompt, $P_t \in\mathbb{R}^{L_e \times D}$ is the e-prompt of $t$-th task, $M$ is the pool size, $L_g$ is the length of g-prompt and $L_e$ is the length of e-prompt.
To retrieve prompts, they set a learnable parameter \( \mathbf{k}_t \in \mathbb{R}^{D} \) for the \( t \)-th e-prompt and form a prompt pool consisting of key-value pairs, \( \mathbf{P} = \{P_g,  (\mathbf{k}_1, P_1), (\mathbf{k}_2, P_2),\cdots, (\mathbf{k}_M, P_M) \} \).
The features extracted by the ViT without prompts, \( f_r(\mathbf{x}_e)[0] \in \mathbb{R}^{D} \), are used as the query \( Q \), and the prompt is selected by:
\begin{equation}
     I=\underset{t\subseteq[1,M]}{\operatorname{argmax}}\cos \langle Q, \mathbf{k}_{t} \rangle,
\end{equation}
where $I$ represents the index of the selected prompt $P_I$ and $\cos \langle \cdot,\cdot\rangle$ denotes cosine similarity function.
Finally, the ViT, along with \( P_I \) and \( P_g \), is used to classify the samples.
Since each sample has only one \( Q \) and each \( P \) corresponds to one \( \mathbf{k} \), we define this query paradigm as the Single Query-Single Key (SQSK) paradigm. 



\subsection{Prompt Matching Rate}
\label{Matching rate}
We define the true prompt as a prompt that was trained on the task to which the sample belongs, i.e., \( I = t \) for a sample \( (\mathbf{x}_i, y_i) \in \mathcal{D}_t \), which means the sample's distribution aligns with the distribution of the prompt's training samples.
Our experiments show that the matching rate under the SQSK paradigm is only 45.15\% in 10-task Split ImageNet-R \cite{hendrycks2021many}.
As shown in \cref{fig:acc_true_false_perfect}, if the samples initially with False Prompt are manually replaced with True Prompt, the performance can be improved by 26.62\% (from 62.06\% to 88.68\%) with the help of the task ID label.
This indicates the significant impact that prompt selection has on performance.
Interestingly, the accuracy of the samples initially with False Prompt, after being replaced with True Prompt, is even higher than that of the samples initially with the True Prompt.
This phenomenon may be attributed to the fact that the tasks associated with samples selected by the False Prompt are more challenging to identify, whereas category prediction within those tasks is relatively easier.
By replacing the False Prompt with the True Prompt, implicit task information is introduced, which leads to improved accuracy—even surpassing that of the samples originally selected by the True Prompt.
Perfect matching refers to an idealized setting where all samples are matched with their corresponding True Prompt. This setting can be viewed as the performance upper bound achievable by prompt selection methods. 
For more detailed settings and explanations of this experiment, please refer to section 1 of supplementary materials.

\section{Method}
Our method uses the same model and tuning approach as SQSK, with the innovation of our method lying in the design of keys and queries, as well as their learning and matching mechanism.
The overall pipeline of our framework is illustrated in \cref{fig:main_picture}.
\subsection{The Design Goal}
We aim to ensure that the test sample follows to the distribution of the training samples used for prompt optimization, as this will enable the prompt to effectively handle the test sample.
To achieve this goal, we need to address the following three questions: 
\begin{enumerate}
    \item How can we extract sample features for similarity judgment?
    \item How can we represent the feature distribution of the samples used for prompt training?
    \item How can we effectively establish a matching mechanism between the features of test samples and the distribution of training samples?
\end{enumerate}

For question 1, we employ a ViT with prompts as the query function.
A ViT with prompts not only incorporates ViT's generalized knowledge but also task-specific fine-grained knowledge, enabling more precise feature extraction for queries.
For question 2, we introduce class-level learnable parameters as keys.
Since samples within a class tend to be similar, class-level keys can represent the feature distribution of a task's samples better.
For question 3, to reduce the interference introduced by queries generated from different prompts, we employ a local matching strategy between each query and the task-specific keys.
With queries, keys and the matching strategy in place, the model can evaluate the similarity between the query and keys, selecting the most consistent prompt for the sample to make the final prediction.


\subsection{Local Matching Mechanism}
We set $|\mathcal{Y}^t|$ different learnable keys $\mathbf{k}_t$ for $P_t$, where $|\mathcal{Y}^t|$ denotes the number of categories in task $t$.
The set of keys $\mathbf{K}_t = \{ \mathbf{k}_{t1}, \mathbf{k}_{t2}, \cdots, \mathbf{k}_{t|\mathcal{Y}^t|} \}$ collectively performs retrieval for $P_t$, where $\mathbf{k}_{tj}$ denotes the key for class $j$ in task $t$.
The keys and prompts together form a key-value pair pool: ${\mathbf{P}=\{P_g, (\mathbf{K}_1, P_1),(\mathbf{K}_2, P_2), \cdots, (\mathbf{K}_M, P_M) \}}$.
We refer to these keys, where a task contains multiple keys for deep search, as MK.
MK was originally introduced by CPrompt \cite{gao2024consistent}, but in fact, our MK differs fundamentally from the MK in CPrompt.
This distinction will be explained in detail in \cref{sec:Optimization_Objective}.

Given a sample $\mathbf{x}$, we use the $t$-th e-prompt and g-prompt in combination with ViT for feature extraction to obtain $f_r(P_g; P_t; \mathbf{x}_e)[0]$, which serves as the $t$-th query $Q_t \in \mathbb{R}^{D}$.
The queries generated by using different e-prompts form a query pool $\mathbf{Q} = \{ Q_1, Q_2, \cdots, Q_M \}$.
We refer to task-level queries for breadth search as MQ.
Next, $Q_t$ is locally matched with $\mathbf{k}_{tj}$ by calculating the cosine similarity $\cos \langle Q_t, \mathbf{k}_{tj} \rangle$.
Since different queries encode knowledge from different tasks, enforcing local matching between each query and its corresponding task allows for more effective utilization of task-specific information and helps mitigate cross-query interference.
At this point, each prompt has matching scores for all the categories within the task.
We need to select the top-$K$ categories with the highest score within a task for aggregation by:\begin{equation}
    \label{eq:aggregated matching score}
    S^{t}=\underset{\{c_m\}_{m=1}^{K}\subseteq[1,|\mathcal{Y}^t|]}{\max}\sum_{m=1}^{K}\cos \langle Q_t, \mathbf{k}_{tc_{m}} \rangle,
\end{equation}
where \( S^t \) is the aggregated matching score between the sample and the \( t \)-th prompt.
Finally, we select the highest matching score, and get the corresponding index as follows:\begin{equation}
    \label{eq:prompt selection}
    I=\underset{t\subseteq[1,M]}{\operatorname{argmax}}\ S^{t}.
\end{equation}
Finally, the model can make predictions according to the selected query $Q_I$ and the linear classifier.

\subsection{Optimization Objective}
\label{sec:Optimization_Objective}
The overall optimization objective consists of two components: (1) performing classification conditioned on the selected prompt, and (2) selecting prompts whose distribution aligns closely with that of the input sample.
Based on the g-prompt and e-prompt, the model is optimized by minimizing the cross-entropy loss between the predicted output and the ground truth label by:\begin{equation}
    \mathcal{L}_\text{cls}=\text{CE}(W^T f_r(P_g;P_t;\mathbf{x}_e)[0],y),
\end{equation}
where $\text{CE}(\cdot,\cdot)$ is cross-entropy function, and $t$ is the task ID.
During the training process, the task ID can be directly used to select the e-prompt.

Since the queries are already guided by the cross-entropy loss in the classification objective, the matching objective focuses solely on aligning the keys with the queries.
Hence, the loss for query-key matching can be written as:\begin{equation}
    \mathcal{L}_{\text{mch}} = \sum_{i=1}^{M} \mathbb{I}(i=t)\sum_{j=1}^{|\mathcal{Y}^i|} (1-\cos\langle Q_i, \mathbf{k}_{ij} \rangle) \mathbb{I}(j=y),
    \label{eq:loss_keys}
\end{equation}
where $\mathbb{I}(\cdot)$ is the indicator function.
The overall loss function can be written as:\begin{equation}
    \mathcal{L} =\mathcal{L}_\text{cls}+\mathcal{L}_\text{mch}.
\end{equation}
Our MK learns by aligning with the MQ that has already been dispersed through cross-entropy.
In contrast, the MK in CPrompt learns by dispersing itself through cross-entropy, since SQ is fixed and not dispersed.
Our MK is a component specifically adapted for MQ.

As shown in \cref{eq:loss_keys}, keys corresponding to other categories and queries corresponding to other prompt are not involved in the loss calculation.
This means that during training, only the designated query is needed, indicating that the MQ mechanism not only avoids introducing additional computational overhead, but actually requires one fewer query without a prompt compared to SQSK.

\subsection{Efficient Inference of MQMK}
MQ integrates knowledge from all tasks for precise querying but comes with a higher inference cost. We aim to design a method that approximates MQ's query accuracy while achieving SQ's query speed, seeking a trade-off between accuracy and speed.
To ensure a speed comparable to SQ, we maintain a non-expanding single-query architecture. However, to incorporate knowledge from all tasks within the query, we need to enhance the query itself. Here, we design a simple yet effective knowledge fusion method to enhance the query by integrating fused knowledge of all tasks. Specifically, we first perform knowledge fusion by averaging all existing e-prompts in the prompt pool and then obtain the enhanced query using the fused prompt and the g-prompt by:
\begin{equation}
    Q^+ = f_r(P_g;\Bar{P};\mathbf{x}_e)[0],
\end{equation}
where $Q^+\in \mathbb{R}^{D}$ is the enhanced query and $\Bar{P}=\frac{1}{t}\sum_{i=1}^{t}P_i$ is the fused prompt.
Finally, we replace \( Q^t \) in MQMK with \( Q^+ \) and select prompts following the same approach as in \cref{eq:aggregated matching score}.
The above describes the Efficient Inference method of MQMK (MQMK-EI). MQMK-EI is an approximation of MQMK designed to accelerate inference. During training, MQMK is more accurate and more efficient. Therefore, MQMK-EI follows the same training process as MQMK, as described in \cref{sec:Optimization_Objective}.

\subsection{The Perspective of Prompt Ensemble}
\label{sec:inference_cost}
MQ can be regarded as a prompt ensemble \cite{lester2021power,schick2020exploiting} query for multiple tasks in continual learning.
Unlike prompt ensemble methods that learn different perspectives of knowledge within a single task, the prompt pool in continual learning stores distinct knowledge across tasks.
In the context of continual learning, directly combining or voting over the logits from different prompts may lead to suboptimal performance.
MQMK employs a multi-query mechanism in which each query is locally aligned with a specific task.
This design helps reduce cross-task interference and knowledge entanglement, making it more suitable for continual learning scenarios characterized by large task discrepancies.

Similar to prompt ensemble methods, we do not need to forward \( M \) times through the ViT backbone to obtain the query pool.
Instead, we replicate a sample \( M \) times, concatenate it with different prompts, and process them in a single batch for parallel inference.
If computational resources are highly limited (e.g., the batch size is restricted to 1), MQMK-EI is more efficient.
If computational resources are abundant, the computation time of MQMK, when parallelized, is reduced to half compared to the two-stage serial computations of MQMK-EI and SQSK.
The MQMK series is training-efficient, achieving approximately a 1/3 speedup compared to the two-stage query process of SQ-based methods.
This is a theoretical analysis, and the practical results, the parameters analysis and computational costs during the training phase, are discussed in section 3 of supplementary materials.


\begin{table*}[t]

\centering
\begin{tabular}{l||cc||cc||cc}
\hline
{Tasks} & \multicolumn{2}{|c||}{5}  & \multicolumn{2}{|c||}{10} & \multicolumn{2}{|c}{20} \\
\hline
Method & $A_T$ (\(\uparrow\)) & $F_T$ (\(\downarrow\)) & $A_T$ (\(\uparrow\)) & $F_T$ (\(\downarrow\)) & $A_T$ (\(\uparrow\)) & $F_T$ (\(\downarrow\)) \\
\hline
joint train & 79.27 & - & 79.27 & - & 79.27 & - \\
\hline
L2P++$^{*}$ & 70.83 $\pm$ 0.58 & 3.36 $\pm$ 0.18 & 69.29 $\pm$ 0.73 & 2.03 $\pm$ 0.19 & 65.89 $\pm$ 1.30 & 1.24 $\pm$ 0.14 \\

Deep L2P++$^{*}$ & 73.93 $\pm$ 0.37 & 2.69 $\pm$ 0.10 & 71.66 $\pm$ 0.64 & 1.78 $\pm$ 0.16 & 68.42 $\pm$ 1.20 & 1.12 $\pm$ 0.13 \\

DualPrompt$^{*}$ & 73.05 $\pm$ 0.50 & 2.64 $\pm$ 0.17 & 71.32 $\pm$ 0.62 & 1.71 $\pm$ 0.24 & 67.87 $\pm$ 1.39 & \underline{1.07 $\pm$ 0.14} \\

ESN$^{\ddagger}$ & 73.42 $\pm$ 0.40 & 3.79 $\pm$ 0.55 & 75.11 $\pm$ 0.36 & 5.68 $\pm$ 0.77 & 70.57 $\pm$ 0.62 & 6.84 $\pm$ 0.36 \\

CODA-P$^{*}$ & 76.51 $\pm$ 0.38 & 2.99 $\pm$ 0.19 & 75.45 $\pm$ 0.56 & \underline{1.64 $\pm$ 0.10} & 72.37 $\pm$ 1.19 & \textbf{0.96 $\pm$ 0.15} \\

EvoPrompt$^{\dagger}$ & 77.16 $\pm$ 0.18 & 9.89 $\pm$ 0.30 & 76.83 $\pm$ 0.08 & 2.78 $\pm$ 0.06 & 74.41 $\pm$ 0.23 & 2.56 $\pm$ 0.22 \\

CPrompt$^{\ddagger}$ & 78.42 $\pm$ 0.14 & 5.15 $\pm$ 0.58 & 77.14 $\pm$ 0.11 & 5.97 $\pm$ 0.68 & 74.79 $\pm$ 0.28 & 7.34 $\pm$ 0.65 \\

\hdashline

MQMK & \textbf{79.61 $\pm$ 0.27} & \underline{1.59 $\pm$ 0.26} & \textbf{78.36 $\pm$ 0.35} & 2.24 $\pm$ 0.20 & \textbf{76.10 $\pm$ 0.17} & 3.33 $\pm$ 0.26 \\
MQMK-EI & \underline{78.50 $\pm$ 0.21} & \textbf{1.24 $\pm$ 0.18} & \underline{78.00 $\pm$ 0.12} & \textbf{1.20 $\pm$ 0.25} & \underline{75.82 $\pm$ 0.18} & 1.87 $\pm$ 0.20 \\
\hline
\end{tabular}

\caption{Results (\%) on Split ImageNet-R under 5-task, 10-task and 20-task settings. Best results are marked in \textbf{bold}. Second-best results are underlined. All our results are over 5 trials. $*$: Results from \cite{smith2023coda}. ${\ddagger}$: Results from \cite{gao2024consistent}. ${\dagger}$: Results from \cite{kurniawan2024evolving}.}
\label{tab:imr}
\end{table*}

\begin{table}[t]

\centering
\begin{tabular}{l||cc}
\hline

Method & $A_T$ (\(\uparrow\)) & $F_T$ \((\downarrow\))\\
\hline
joint train & 91.79 & - \\
\hline
L2P++$^{*}$ & 82.50 $\pm$ 1.10 & 1.75 $\pm$ 0.42  \\

Deep L2P++$^{*}$ & 84.30 $\pm$ 1.03 & \textbf{1.53 $\pm$ 0.40} \\

DualPrompt$^{*}$ & 83.05 $\pm$ 1.16 & 1.72 $\pm$ 0.40 \\

ESN$^{\ddagger}$ & 86.42$\pm$ 0.80 & 6.08$\pm$ 0.48 \\

CODA-P$^{*}$ & 86.25 $\pm$ 0.74 & {1.67 $\pm$ 0.26}  \\

EvoPrompt$^{\dagger}$ & 87.97 $\pm$ 0.30 &2.60 $\pm$ 0.42 \\

CPrompt$^{\ddagger}$ & 87.82 $\pm$ 0.21 & 5.06 $\pm$ 0.50 \\

\hdashline
MQMK & \underline{91.73 $\pm$ 0.18} & {2.67 $\pm$ 0.17}  \\
MQMK-EI & \textbf{92.00 $\pm$ 0.29} & \underline{1.58 $\pm$ 0.18}  \\
\hline
\end{tabular}

\caption{Results (\%) on Split CIFAR-100 under 10-task setting. Best results are marked in \textbf{bold}. Second-best results are underlined. All our results are over 5 trials. $*$: Results from \cite{smith2023coda}. ${\ddagger}$: Results from \cite{gao2024consistent}. ${\dagger}$: Results from \cite{kurniawan2024evolving}.}
\label{tab:cifar}
\end{table}

\section{Experiments}

\subsection{Evaluation Benchmarks}
\textbf{Dataset.}
We shuffle the classes and perform task splitting on three widely used visual datasets for prompt-based continual learning: CIFAR-100 \cite{krizhevsky2009learning}, ImageNet-R \cite{hendrycks2021many}, and DomainNet \cite{peng2019moment}, in order to align with the problem setup and conduct comprehensive experiments.
We divide CIFAR-100 and DomainNet into 10 tasks, while ImageNet is divided into three cases: 5 tasks, 10 tasks, and 20 tasks.
The detailed dataset introduction is in section 8 in supplementary materials.
The data distributions of DomainNet and ImageNet-R differ significantly from the ImageNet dataset used for model pre-training.
This requires the model to continually learn new knowledge from these datasets, rather than relying on the knowledge learned from the pre-training dataset.

\textbf{Evaluation Metrics.}
Average accuracy \cite{kim2023stability,yan2021dynamically,li2017learning} and forgetting rate \cite{chaudhry2018efficient,lopez2017gradient} are the two core metrics we use.
$A_T$ is the average accuracy on tasks 1 to $T$ after the model has learned task $T$, and can be computed by:
\begin{equation}
    A_T = \frac{1}{T} \sum_{t=1}^{T} \text{Accuracy}(t,T),
\end{equation}
where $\text{Accuracy}(t,T)$ represents the accuracy on task $t$ after learning task $T$.
$F_T$ is the average accuracy drop across all tasks, and can be computed by:
\begin{equation}
    F_T = \frac{1}{T} \sum_{t=1}^{T} \left( \text{Accuracy}\left(t,t\right) - \text{Accuracy}\left(t,T\right) \right).
\end{equation}
Average accuracy directly reflects the overall performance of the model, while forgetting rate indicates the trend of performance change.
Therefore, average accuracy is relatively more important.

\textbf{Implementation Details.}
ViT-B/16 \cite{dosovitskiy2020image} pre-trained on ImageNet-21K \cite{deng2009imagenet} and fine-tuned on ImageNet-1K \cite{russakovsky2015imagenet} is the backbone we use, so $S$ is set to 16.
G-prompt is used in the first two layers of MSA, with a length of 5.
The depth and length of e-prompt are discussed in \cref{sec:len_depth}.
In all experiments, \( M \) is set equal to \( T \), meaning that only one prompt is learned for each task.
Each prompt selects only the category with the highest similarity for aggregation which implies that \( K \) is set to 1.
For both SQSK and MQMK, a learning rate of 0.005, a batch size of 64, and the Adam \cite{kingma2014adam} optimizer with \( \beta_1 = 0.9 \) and \( \beta_2 = 0.999 \) are used in all experiments.

\begin{table}[t]

\centering
\begin{tabular}{l||cc}
\hline

Method & $A_T$ (\(\uparrow\)) & $F_T$ \((\downarrow\))\\
\hline
joint train & 89.15 & - \\
\hline
L2P & 81.17 $\pm$ 0.83 & 8.98 $\pm$ 1.25  \\

DualPrompt & 81.70 $\pm$ 0.78 & 8.04 $\pm$ 0.31 \\

ESN & 79.22 $\pm$ 2.04 &10.62$\pm$ 2.12 \\

CODA-P & 80.04 $\pm$ 0.79 & {10.16 $\pm$ 0.35}  \\

CPrompt & 82.97 $\pm$ 0.34 & 7.45 $\pm$ 0.93 \\

\hdashline
MQMK & \textbf{85.62 $\pm$ 0.33} & \underline{5.51 $\pm$ 0.22}  \\
MQMK-EI & \underline{84.54 $\pm$ 0.36} & \textbf{3.33 $\pm$ 0.40}  \\
\hline
\end{tabular}
\caption{Results (\%) on Split DomainNet under 10-task setting. Best results are marked in \textbf{bold}. Second-best results are underlined. All our results are over 5 trials. Except for MQMK, all the other results come from \cite{gao2024consistent}.}
\label{tab:domainnet}
\end{table}

\textbf{Competitors and Joint Training.}
We compare our method with SOTA prompt-based continual learning models, including L2P \cite{wang2022learning}, DualPrompt \cite{wang2022dualprompt}, ESN \cite{wang2023isolation}, CODA-P \cite{smith2023coda}, EvoPrompt \cite{kurniawan2024evolving} and CPrompt \cite{gao2024consistent}.
To ensure a fair comparison, we use SQSK as a baseline in some experiments.
SQSK and MQMK have identical prompts, with the only difference being the matching mechanism.
SQSK can be seen as a version of DualPrompt with adjusted hyper-parameters, and it has nearly the same performance as DualPrompt reported in \cref{tab:imr,tab:cifar,tab:domainnet}.
Additionally, we jointly train a model using data from all tasks with linear prototypes and prompt tuning as a reference for the upper bound.

\subsection{Performance Comparison}
As shown in \cref{tab:imr}, in the three task settings on ImageNet-R, MQMK achieves SOTA performance. In the 5-task and 10-task splits, MQMK shows no significant performance gap compared to joint training.
For the small task split, the continual learning performance with MQMK no longer shows a significant gap compared to traditional joint training with all data.
This indicates that the performance on ImageNet-R with the 5-task and 10-task splits is relatively close to its oracle.
However, on the 20-task split, MQMK still shows a noticeable performance gap compared to joint training.
A possible explanation is that as the number of tasks increases, the limited data per task becomes inadequate for training high-quality e-prompts, ultimately resulting in a degradation of overall performance.
Additionally, compared to the high-performing CPrompt and EvoPrompt, MQMK has a clear advantage in terms of forgetting rate.
This indicates that MQMK not only achieves higher final performance, but also experiences smaller performance degradation.
CODA-P has a low forgetting rate, yet its average accuracy is also relatively low.
This suggests that CODA-P exhibits overall stability in performance, but its final performance is average.

On CIFAR-100, \textbf{MQMK also achieves performance close to that of joint training, improving by 3.91\% compared to CPrompt}, while maintaining a low forgetting rate.
On DomainNet, MQMK shows a 2.65\% performance improvement over CPrompt, along with the lowest forgetting rate.
Split-DomainNet is an class-imbalanced dataset, where MQMK achieves 90\% accuracy on some tasks, while on tasks with small sample sizes, the accuracy drops to only 60\%.
The issues of small samples and class imbalance may be the reasons for the poor performance of all continual learning algorithms.

Overall, MQMK-EI slightly lags behind MQMK but still achieves SOTA performance. 
On CIFAR-100 and the 20-task split of ImageNet-R, MQMK-EI achieves performance that is very close to MQMK. This is because CIFAR-100 is a relatively simple dataset with consistent image styles, where knowledge fusion among different prompts introduces minimal loss or interference. In the 20-task split of ImageNet-R, the limited number of samples per task makes it difficult to train high-quality e-prompts. In such cases, knowledge fusion helps to enhance prompt quality to a certain extent.

\begin{figure}[t]
  \centering
  \begin{subfigure}{0.48\linewidth}
    \includegraphics[width=\textwidth, trim=0 10 0 10 , clip]{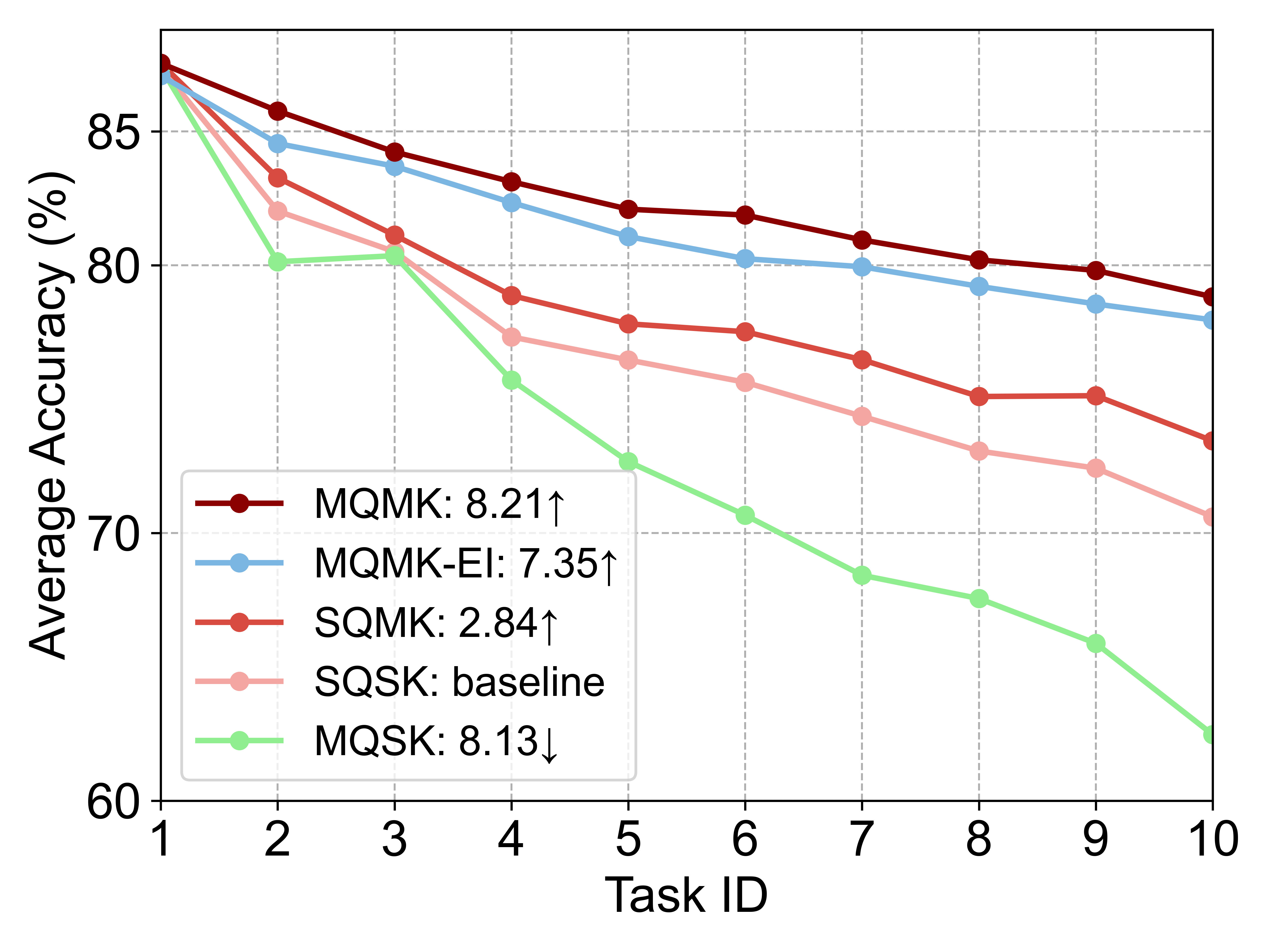}
    \caption{Average Accuracy.}
  \end{subfigure}
  \begin{subfigure}{0.48\linewidth}
    \includegraphics[width=\textwidth, trim=0 10 0 10, clip]{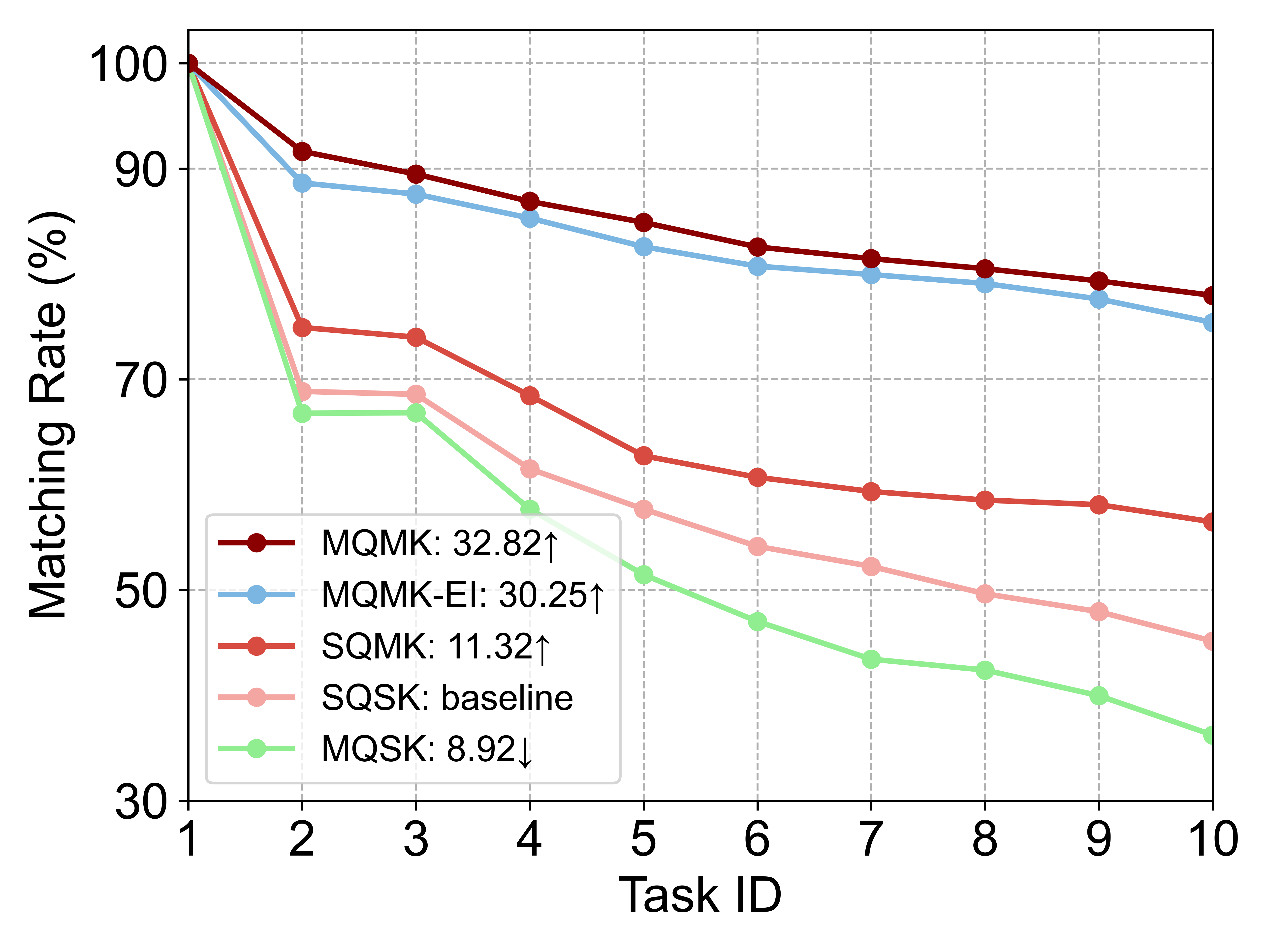}
    \caption{Matching Rate.}
  \end{subfigure}
  \caption{Ablation experiments on the 10-task Split ImageNet-R. The average accuracy and matching rate change with the learning process.}
  \label{fig:ablation}
\end{figure}

\begin{table}[t]

\centering
\begin{tabular}{c||cc}
\hline
Number & $A_T$ (\(\uparrow\)) & Matching Rate \((\uparrow\))\\
\hline
1 & 62.47 & 36.23 \\

4 & 71.20  & 57.82  \\

10 & 75.17  & 68.77 \\

20 & 78.82 & 77.97 \\

\hline
\end{tabular}
\caption{$A_T$ (\%) and matching rate (\%) under keys of different granularities on 10-task Split Imagenet-R.}
\label{tab:number_key}
\end{table}

\begin{figure}
  \centering
  \begin{subfigure}{0.23\textwidth}
    \includegraphics[width=\linewidth, trim=0 0 0 0, clip]{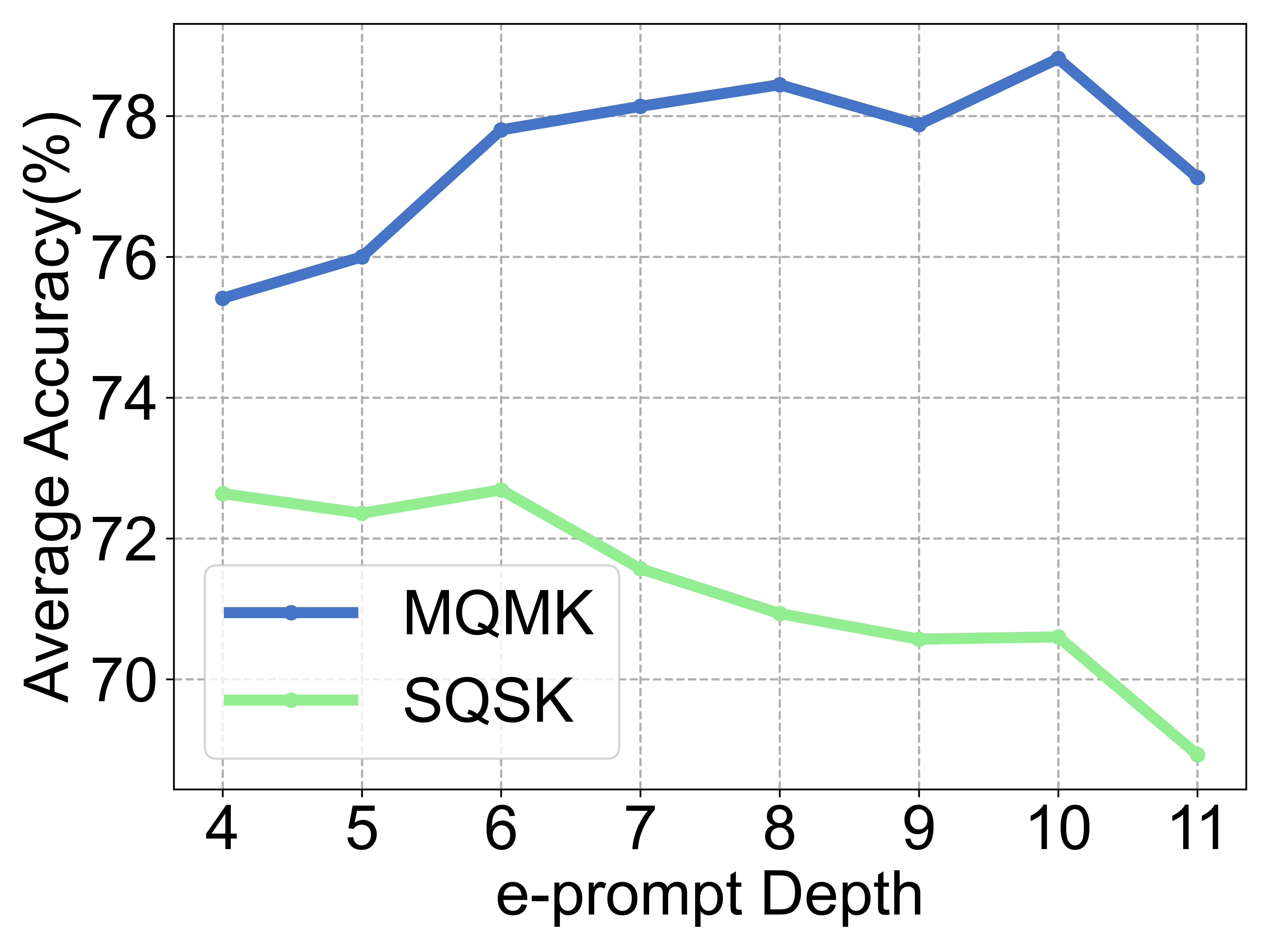}
    \caption{Different depth.}
    
  \end{subfigure}
  \begin{subfigure}{0.23\textwidth}
    \includegraphics[width=\linewidth, trim=0 0 0 0, clip]{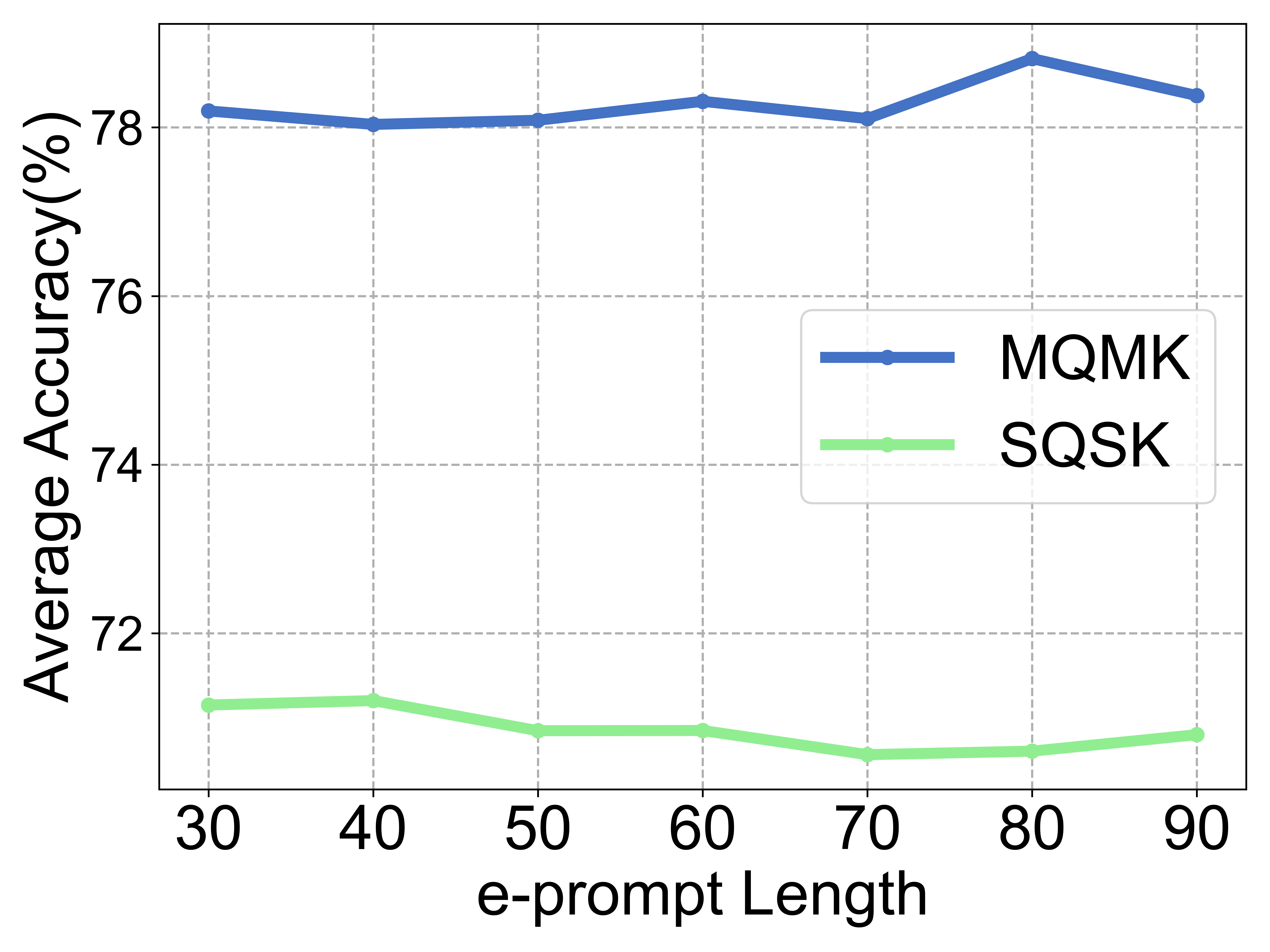}
    \caption{Different length.}
   
  \end{subfigure}
  \caption{Average accuracy of MQMK and SQSK on the 10-task Split ImageNet-R under different depths and lengths of the e-prompt. (a) The length is fixed at 80, and the depth varies. (b) The depth is fixed at 10, and the length varies.}
  \label{fig:depth_length}
\end{figure}

\subsection{Ablation Study}
The components of our method include Multiple Queries (MQ) and Multiple Keys (MK).
The corresponding components in existing methods are Single Query (SQ) and Single Key (SK).
We discuss the average accuracy and matching rate under four combinations: MQMK, MQSK, SQMK, and SQSK, as well as the inference-efficient variant MQMK-EI, as illustrated in \cref{fig:ablation}.
Using SQSK as the baseline and adding MK, $A_T$ improves by 2.84\% and the matching rate increases by 11.32\%.
This demonstrates the effectiveness of setting class-level keys for each prompt.
By adding MQ, $A_T$ decreases by 8.13\% and the matching rate drops by 8.92\%.
This is because, under the supervision of cross-entropy, the features of samples from the same class are more clustered, while the features of samples from different classes are more dispersed.
The features of multiple classes in a task have already been dispersed, and in this case, SK can no longer represent the feature distribution of all training samples in a task.
\textbf{When MQ is combined with MK, $A_T$ increases by {8.21\%} and the matching rate improves by 32.82\%}.
This demonstrates that incorporating task-related knowledge significantly enhances retrieval accuracy.
In fact, in query-key matching, the improvement of query quality and key quality are complementary.
Enhancing both simultaneously has a significant impact on matching rate.
MQMK-EI achieves performance comparable to MQMK while incurring a lower query cost.

\subsection{Setting Fine-grained Keys}
MQSK leads to a performance decline, whereas MQMK improves performance.
This highlights the critical importance of learning reliable keys for distribution representation.
Based on MQ, we gradually increase the number of keys from 1 to the number of classes in each task (20 class per task in 10-task Split Imagenet-R), and we report their performance in \cref{tab:number_key}.
We divide each task into several class groups based on the number of keys, with each group containing multiple classes.
From this perspective, both SK and MK are special cases of $N$-keys.

\begin{table}[ht]
\centering
\begin{tabular}{l||ccc}
\hline
 Method & DomainNet & ImageNet-R & CIFAR-100 \\
\hline
L2P          & 72.31 & 42.82 & 48.73 \\
ESN          & 72.84 & 70.33 & 78.50 \\
CPrompt      & 75.36 & 64.00 & 69.37 \\
SQSK         & 70.55 & 45.15 & 49.14 \\
\hdashline
MQMK         & 82.24 & 77.97 & 82.28 \\
MQMK-EI      & 76.67 & 75.40 & 86.17 \\
\hline
\end{tabular}
\caption{Matching rate (\%) of different methods on three 10-task split benchmarks.}
\label{tab:Matching_Rates}
\end{table}

It can be observed that \textbf{as the number of keys increases, both the matching rate and accuracy improve}, and the optimal performance is achieved when the key reaches the class level.
Since the smallest granularity of our labels is at the class level, when the number of keys exceeds the number of classes, it becomes difficult to provide appropriate supervisory signals for the keys.

\begin{figure}[t]
  \centering
  \begin{subfigure}{0.48\linewidth}
    \includegraphics[width=\textwidth, trim=0 0 0 0, clip]{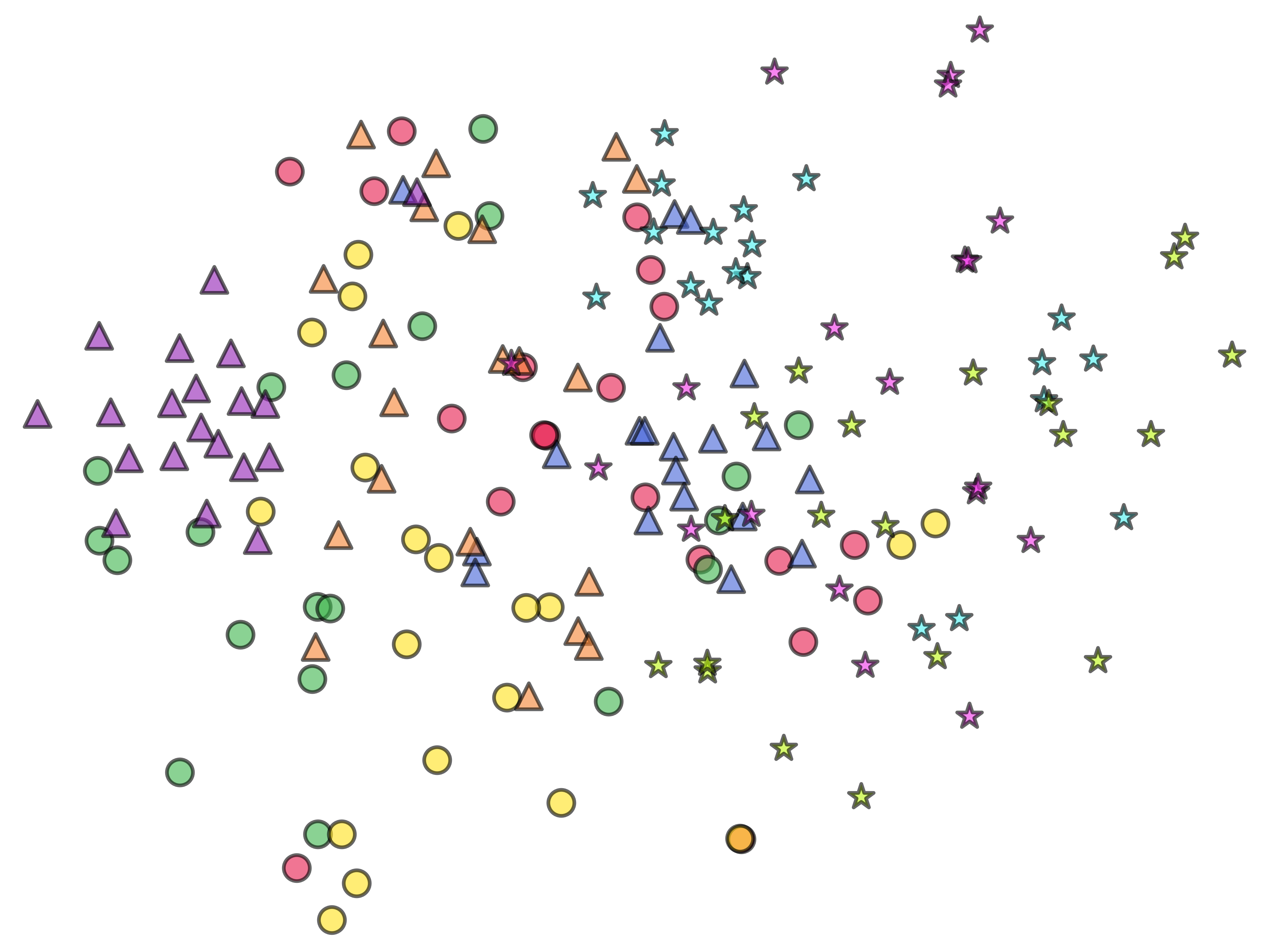}
    \caption{SQ.}
  \end{subfigure}
  \begin{subfigure}{0.48\linewidth}
    \includegraphics[width=\textwidth, trim=0 0 0 0 , clip]{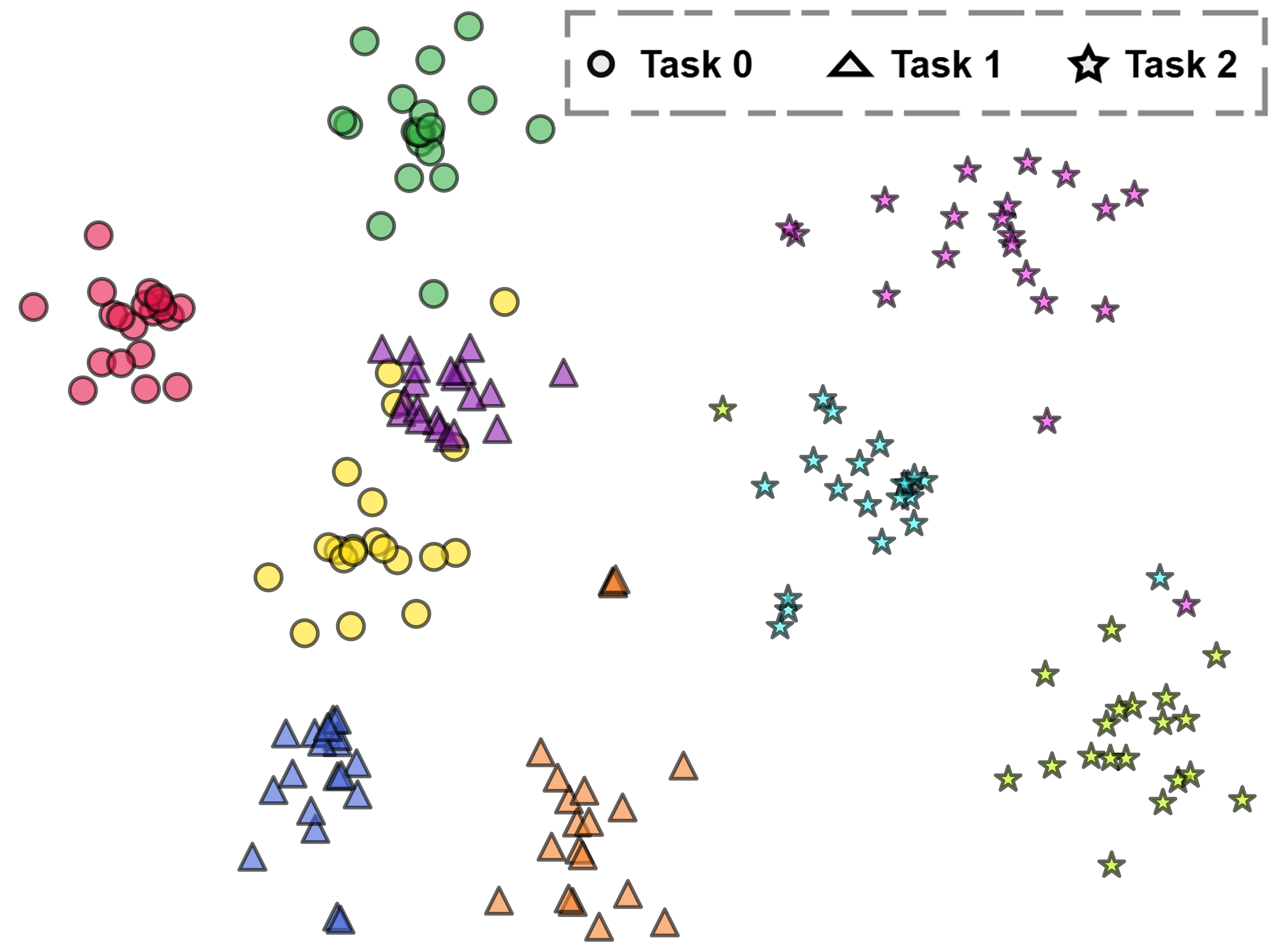}
    \caption{Selected Queries in MQ.}
  \end{subfigure}
  
  \caption{t-SNE \cite{van2008visualizing} visualization of the Queries. The samples come from three classes of three tasks in the 5-task Split Imagenet-R. The shapes represent the sample tasks, and the colors represent the sample categories.}
  \label{fig:tsne}
\end{figure}

\subsection{Going Deeper and Longer}
\label{sec:len_depth}
As shown in \cref{fig:depth_length}, the performance of the MQMK method improves as the e-prompt depth and length increase, reaching its optimal performance of 78.82\% when the e-prompt depth is 10 and length is 80.
In contrast, the performance of the SQSK method decreases as the prompts go deeper and longer.
A possible reason is that increasing the depth and length of the prompt can improve its quality.
The higher matching rate of MQMK allows it to effectively utilize the higher-quality prompt, thereby enhancing performance.
In contrast, SQSK does not benefit in the same way.
This suggests that \textbf{an accurate query paradigm can allow the prompt to go deeper and longer, thereby improving model performance}.

\subsection{Visualization of Queries}
\label{sec:visualization}
We present visualizations of the queries selected by MQ, as well as queries in SQ, as shown in \cref{fig:tsne}.
It can be observed that the queries selected by MQ exhibit stronger intra-class cohesion and inter-class separation, indicating that MQ queries are precise and the introduction of prompts is necessary.
On the contrary, due to SQ's reliance solely on ViT's generalization ability and the lack of task-specific knowledge, the queries of SQ are highly dispersed, resulting in almost failed clustering.
Additionally, although the queries show a tendency to cluster by task, the smallest granularity of clustering is at the class level, which indicates the necessity of setting class-level keys.

\begin{figure}[t]
  \centering
  \begin{subfigure}{0.23\textwidth}
    \includegraphics[width=\linewidth, trim=0 0 0 0, clip]{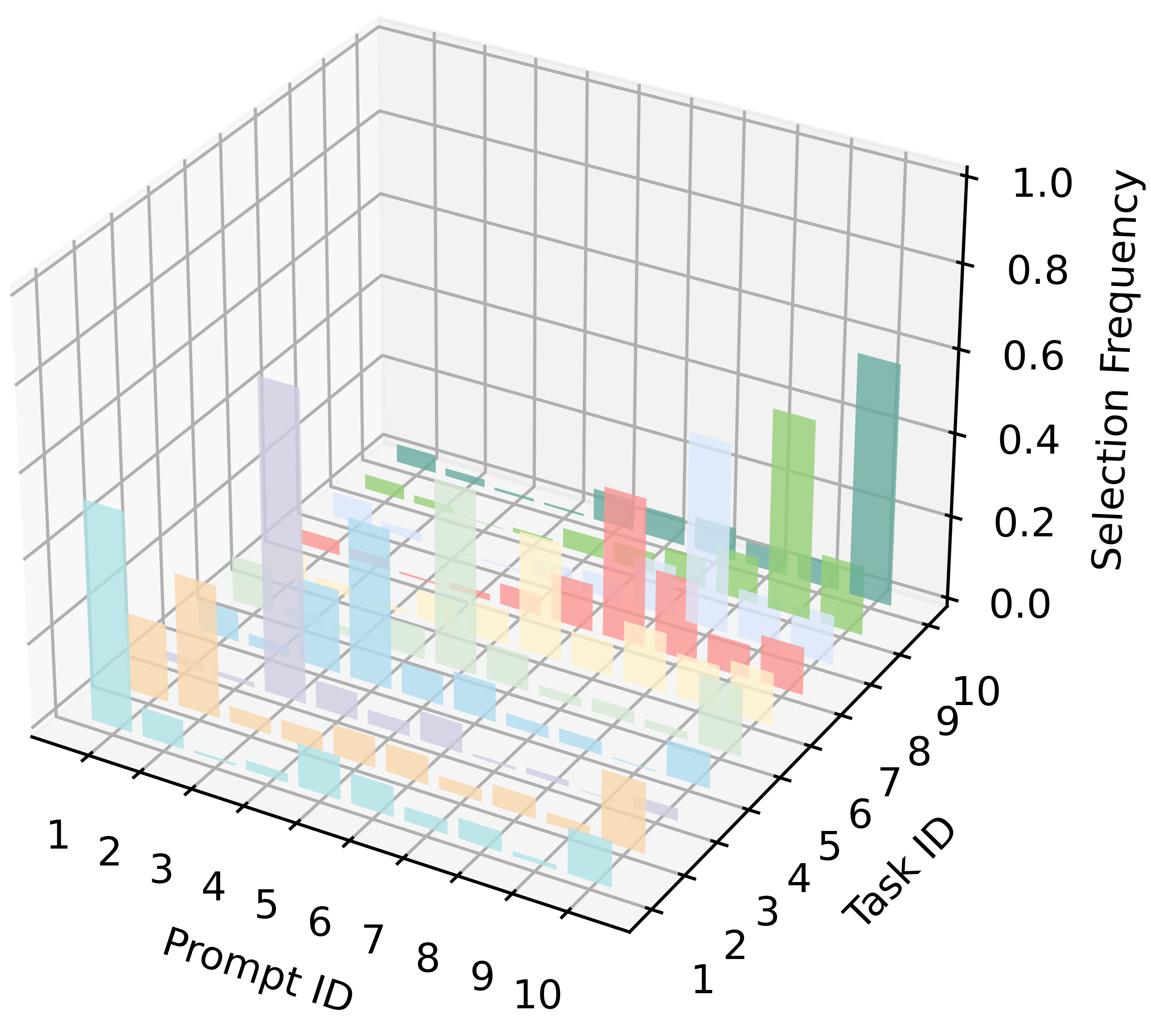}
    \caption{SQSK.}
    
  \end{subfigure}
  \begin{subfigure}{0.23\textwidth}
    \includegraphics[width=\linewidth, trim=0 0 0 0, clip]{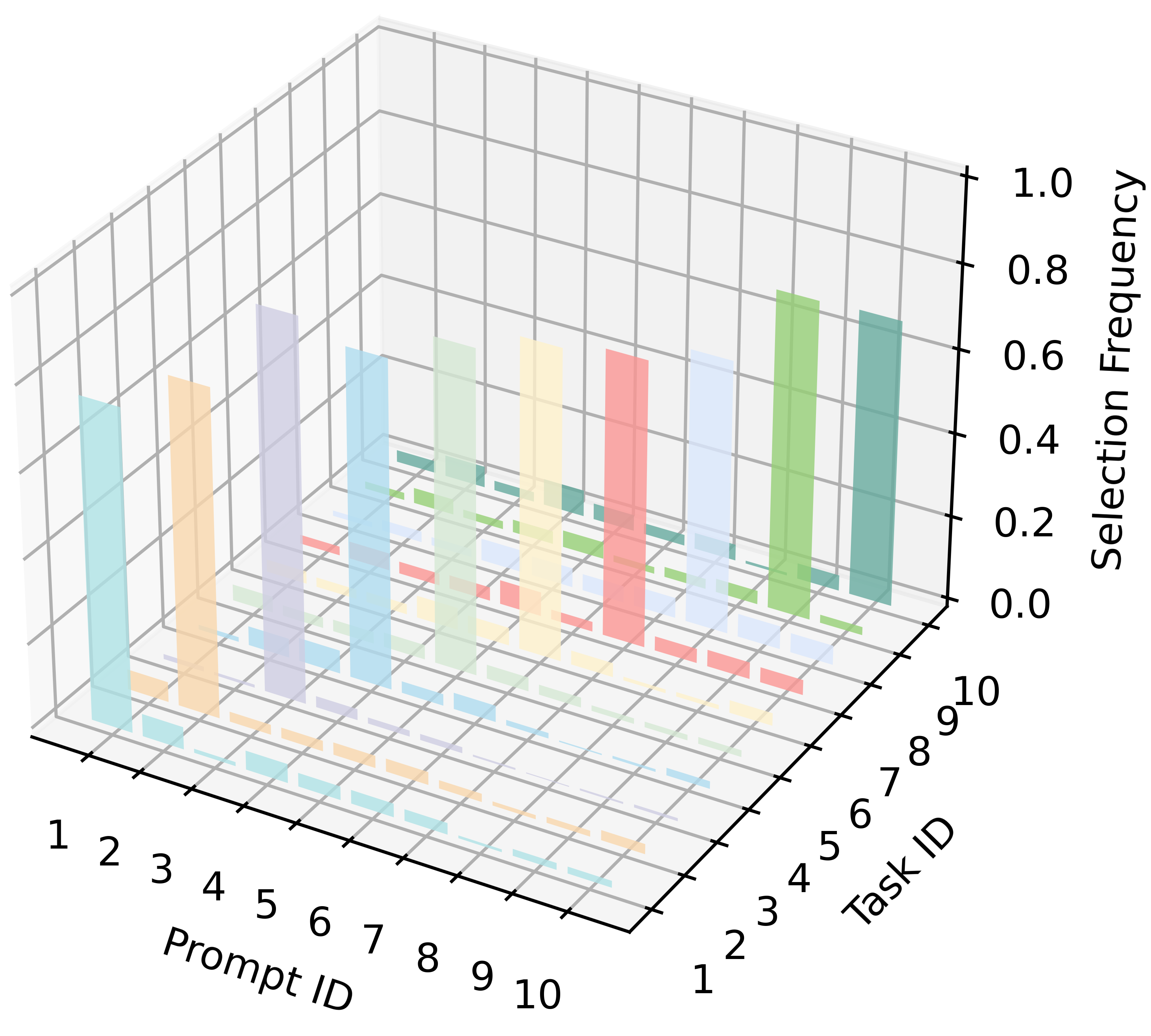}
    \caption{MQMK.}
   
  \end{subfigure}
  \caption{The 3D visualization of specific prompt selection process of MQMK and SQSK on 10-task Split Imagenet-R. }
  \label{fig:3d_visualize}
\end{figure}

\section{Matching Rate Across Methods}
As shown in \cref{tab:Matching_Rates}, MQMK improves the matching rate by over 30\% compared to SQSK on the challenging CIFAR-100 and ImageNet-R.
Due to the significant differences in data styles across different domains, SQSK can achieve high performance on DomainNet.
To gain a deeper understanding of the prompt selection mechanism, we visualize the specific prompt selection process of MQMK and SQSK, generating two 3D bar charts, as shown in \cref{fig:3d_visualize}.
The results indicate that MQMK exhibits a significantly higher probability of selecting the true prompt compared to SQSK.
MQMK outperforms MQMK-EI on datasets with large style discrepancies, such as ImageNet-R and DomainNet, whereas on more consistent datasets like CIFAR-100, MQMK-EI, with its knowledge fusion enhanced queries, achieves higher matching rates.
Both experiments demonstrate the effectiveness of our approach in improving the matching rate.

CODA-P and EvoPrompt use soft, weighted prompt selection, where correctness does not exist. The matching objective of ESN is task-specific classifiers, which bears similarity to prompt selection. According to ablation studies, MQMK is identical to other variants such as SQSK in all aspects except for the prompt strategy. In vertical comparisons with other methods, MQMK still demonstrates a clear advantage in terms of matching rate.

\section{Conclusion}
To address the issue of low prompt selection accuracy in prompt-based continual learning, we propose the Multiple Queries with Multiple Keys prompt local matching paradigm.
Multiple Queries achieve precise feature extraction by incorporating task-specific knowledge.
Multiple Keys, through fine-grained learnable keys, better represent the feature distribution of the training samples.
They complement each other, significantly improving the matching rate and performance through local matching mechanism.
An accurate parameter selection strategy can bring significant performance improvements to existing continual learning methods based on parameter expansion and selection. Although precise querying incurs a certain query cost, MQMK-EI maintains high query accuracy while effectively reducing the query cost.

\begin{acks}
This work was partially supported by the STI 2030-Major Projects of China (Grant No. 2021ZD0201300), the National Natural Science Foundation of China (Grant Nos. 62276127, 62495094), and the Fundamental Research Funds for the Central Universities (Grant No. 2024300394). The authors gratefully acknowledge these supports.
\end{acks}

\bibliographystyle{ACM-Reference-Format}
\bibliography{acmart}


\newpage





\title{Supplementary Materials: Multiple Queries with Multiple Keys: A Precise Prompt Matching Paradigm for Prompt-based Continual Learning}


\maketitle
\setcounter{section}{0} 
\section{Prompt Matching Rate}
\label{Matching rate}

We begin by formally defining a \emph{True Prompt}: for any test sample \((\mathbf{x}_i, y_i)\) drawn from task \(t\), a prompt \(p_j\) is considered \emph{true} if and only if it was trained on task \(t\), i.e., the sample's data distribution matches the prompt's training distribution (\(I = t\)).

Under the standard Single Query--Single Key (SQSK) paradigm, our experiments on Split ImageNet-R (10 tasks, 6{,}000 test samples) show that only 45.15\% of samples are paired with their True Prompt.

To better analyze the impact of prompt selection errors, we divide the test set into two disjoint subsets based on whether SQSK correctly matches the True Prompt:
\begin{itemize}
    \item \textbf{False Prompt}: 1{,}232 samples for which SQSK selected an incorrect prompt.
    \item \textbf{True Prompt}: 4{,}768 samples for which SQSK selected the correct prompt.
\end{itemize}

We then evaluate four experimental settings, shown in Figure 2 of the main text.

\begin{itemize}
    \item \textbf{False Prompt (62.06\%)}: Accuracy on the ``False Prompt'' subset when using the initially retrieved (incorrect) prompts.
    \item \textbf{True Prompt (79.33\%)}: Accuracy on the ``True Prompt'' subset with correctly retrieved prompts.
    \item \textbf{False $\rightarrow$ True (88.68\%)}: Accuracy on the same 1{,}232 ``False Prompt'' samples, after manually replacing the incorrect prompts with their corresponding True Prompts. Since the sample set remains unchanged, this directly measures the performance gain from correcting prompt selection.
    \item \textbf{Perfect Match (84.46\%)}: Accuracy on the entire 6{,}000 test samples when each sample is matched with its True Prompt, representing an ideal upper bound.
\end{itemize}

Several important observations emerge from these results:
\begin{enumerate}
    \item The comparison between ''False Prompt'' and ``False $\rightarrow$ True'' is fair because they use exactly the same sample set; the large accuracy gain (+26.62\%) demonstrates the significant negative impact of incorrect prompt selection.
    \item Interestingly, ``False $\rightarrow$ True'' achieves even higher accuracy than the ``True Prompt'' subset. This may be because the samples initially mismatched by SQSK tend to belong to tasks that are harder to identify, but once the correct prompt is provided, these samples become relatively easier to classify within their task.
    \item "Perfect Match" represents the upper bound that could be achieved by an ideal prompt selection mechanism, confirming the importance of improving the prompt matching rate.
    It should be noted that perfect matching involves information leakage, so its accuracy cannot be used as a baseline for performance comparison, but only as a reference for the upper bound of performance.

\end{enumerate}

In summary, these experiments clearly show that prompt selection is a critical factor affecting performance under SQSK. Motivated by this, we propose the Multiple Queries with Multiple Keys (MQMK) paradigm to improve the prompt matching rate and thus boost overall accuracy.

\section{Key Matching Classifier}
\begin{table}[ht]

\centering
\begin{tabular}{c||ccc}
\hline
Classifier & FC & NCM & KM\\
\hline
$A_T$  & 78.82 & 77.25 & 72.27\\

\hline
\end{tabular}
\caption{$A_T$ (\%) of MQMK using 3 different classifiers on 10-task Split Imagenet-R.}
\label{tab:different classifiers}
\end{table}
When $K=1$, our method can be interpreted as selecting the key with the highest cosine similarity, and choosing the corresponding prompt.
Since MK extends keys to the class level, establishing an identity mapping between keys and classes and directly deriving the classification result upon key selection is more straightforward and worth exploring.
We refer to this classifier, which directly classifies based on the selected class-level keys, as the Key Matching (KM) classifier.

We compare the performance of the KM classifier with the Fully Connected (FC) classifier used throughout the experiment and an additional classic prototype-based Nearest Class Mean (NCM) classifier.
As shown in \cref{tab:different classifiers}, the performance of the KM classifier was not satisfactory, failing to surpass both the FC and NCM classifiers.
Notably, on the 10-task Split ImageNet-R, the matching rate reached 77.97\%.
Since the classification accuracy of the KM classifier essentially reflects the accuracy of key matching, while the matching rate represents the accuracy of prompt selection, the gap between them (5.70\%) corresponds to the samples where key matching errors occur but the prompt is correctly selected.
\textbf{The matching rate can be considered a loose upper bound of the KM classifier’s accuracy.}
In this task, the matching rate is lower than the FC classifier’s accuracy, which explains why the KM classifier fails to outperform the FC classifier.
However, if the matching rate were to improve further, it remains to be explored whether the KM classifier could achieve performance comparable to or even surpassing that of the FC classifier.



\section{Discussion on Computational Costs and Number of Parameters}
\label{sec:cost_analysis}

\begin{table*}[ht]
\centering
\begin{tabular}{l||cccc}
\hline
Method & Training (h) & Inference (ms) & Learnable Parameters (M) & All Parameters (M)\\
\hline

SQSK & 4.28 & 14.65  & 5.63 & 87.88\\

MQMK & 2.78 &  72.62  & 5.70  & 87.95\\

MQMK (Parallel) & 2.78 &  7.48  & 5.70  & 87.95\\

MQMK-EI & 2.78 &  14.86  & 5.70  & 87.95 \\

\hline
\end{tabular}
\caption{Computational costs and number of parameters for SQSK and MQMK with $L_e = 40$, $H_e = 7$, $L_g = 5$, and $H_g = 2$ on the 10-task split of CIFAR-100, evaluated on a single NVIDIA RTX 4090 GPU.}
\label{tab:cost_analysis}
\end{table*}
We have discussed the theoretical cost of MQMK.
In this section, we will discuss the practical computational costs, theoretical training computational costs, and the number of parameters.

\textbf{Number of Trainable Parameters.}
The total number of parameters in the prompt is \( L_g \times H_g \times D + L_e \times H_e \times D \), where $H_g$ represents the number of layers inserted by the g-prompt and $H_e$ represents the number of layers inserted by the e-prompt.
The number of parameters for SK is \( M \times D \), and the number of parameters for MK is \( |\mathcal{Y}^1 \cup \mathcal{Y}^2 \cup \cdots \cup \mathcal{Y}^T| \times D \).
Compared to SQSK, the prompt and classifiers parameters of MQMK remain unchanged, while the key parameters become multiple times the number of categories in each task.


\textbf{Training Process.} Queries are directly located based on the task $t$.
Since the prediction and the query use the same features, only one forward and one backward pass through the ViT backbone is required.
In contrast, SQSK needs to perform both a query and a prediction with the prompt, which requires two forward passes and one backward pass through the ViT backbone.

\textbf{Practical Results.} Training Phase: The MQMK series is more efficient, saving approximately 1/3 of the training time. This is because SQ-based methods require running the ViT backbone twice—once for querying and once for inference using the selected prompt. In contrast, MQMK performs both querying and inference based on the same feature representation, avoiding redundant computation.
Inference Phase: Both MQMK-EI and SQSK employ a two-stage sequential inference process, leading to comparable inference times. However, MQMK (Parallel) significantly improves inference efficiency by replicating each input $M$ times, combining them with different prompts, and forwarding them as a single batch of size $M$. This parallelization strategy results in nearly 2× speedup.
While the sequential inference variant of MQMK incurs higher latency than other approaches, its speed is still acceptable in many real-world applications.
The practical results align with the theoretical expectations.

\section{Visualization of Features in Perfect Match}
\begin{figure}[ht]
    \centering
    \includegraphics[width=0.4\textwidth,trim=0 0 0 0,clip]{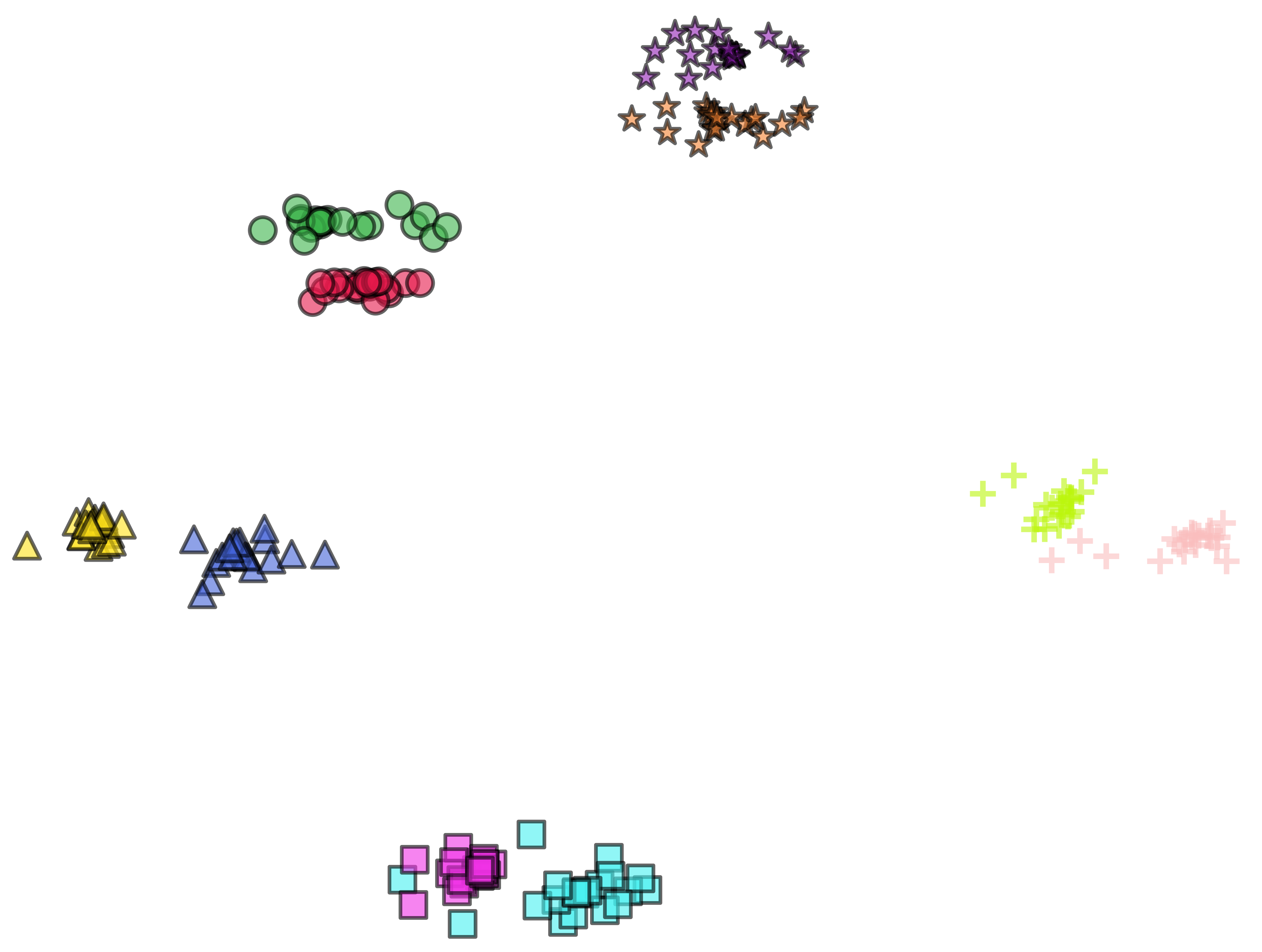}
    \caption{t-SNE \cite{van2008visualizing} visualization of features in perfect match. The samples come from two classes across five tasks in the 5-task Split Imagenet-R. The shapes represent the sample tasks, and the colors represent the sample categories.
    }
    \label{fig:tsne_perfectmatch}
\end{figure}
We have discussed the query visualization of MQ and SQ.
In this section, we provide feature visualizations under the condition where the model selects the true prompt for all samples (perfect match), which can be considered an ideal upper bound.
As shown in \cref{fig:tsne_perfectmatch}, when the true prompt is selected, the discriminability of the features is very strong.
Therefore, improving prompt matching rate can significantly enhance performance.
Additionally, there is no inherent relationship between classes within the same task.
\textbf{However, since the same prompt is selected, the features of samples in the same task become very similar.
It suggests that different prompts generate different subspaces, and selecting the wrong prompt may cause subspace drift, leading to biased predictions.}
This highlights that the selection of the prompt is crucial.

\begin{table*}[ht]
\centering
\begin{tabular}{l||ccccccccc}
\hline
Method & 2 & 3 & 4 & 5 & 6 & 7 & 8 & 9 & 10 \\
\hline
MQMK &86.53	&86.27 &86.13& 85.73& 85.20& 84.53&	84.40&	84.40&	83.73\\

MQMK-EI & 84.93&	84.80&	84.93&	84.33&	84.66&	83.73&	83.73&	82.67&	83.33\\

\hline
\end{tabular}
\caption{Accuracy of task 2 across time, on 10-task Split Imagenet-R}
\label{tab:task_2_Accuracy}
\end{table*}

\section{Visualization of Queries in MQMK-EI}
\begin{figure}[t]
  \centering
  \begin{subfigure}{0.48\linewidth}
    \includegraphics[width=\textwidth, trim=0 0 0 0, clip]{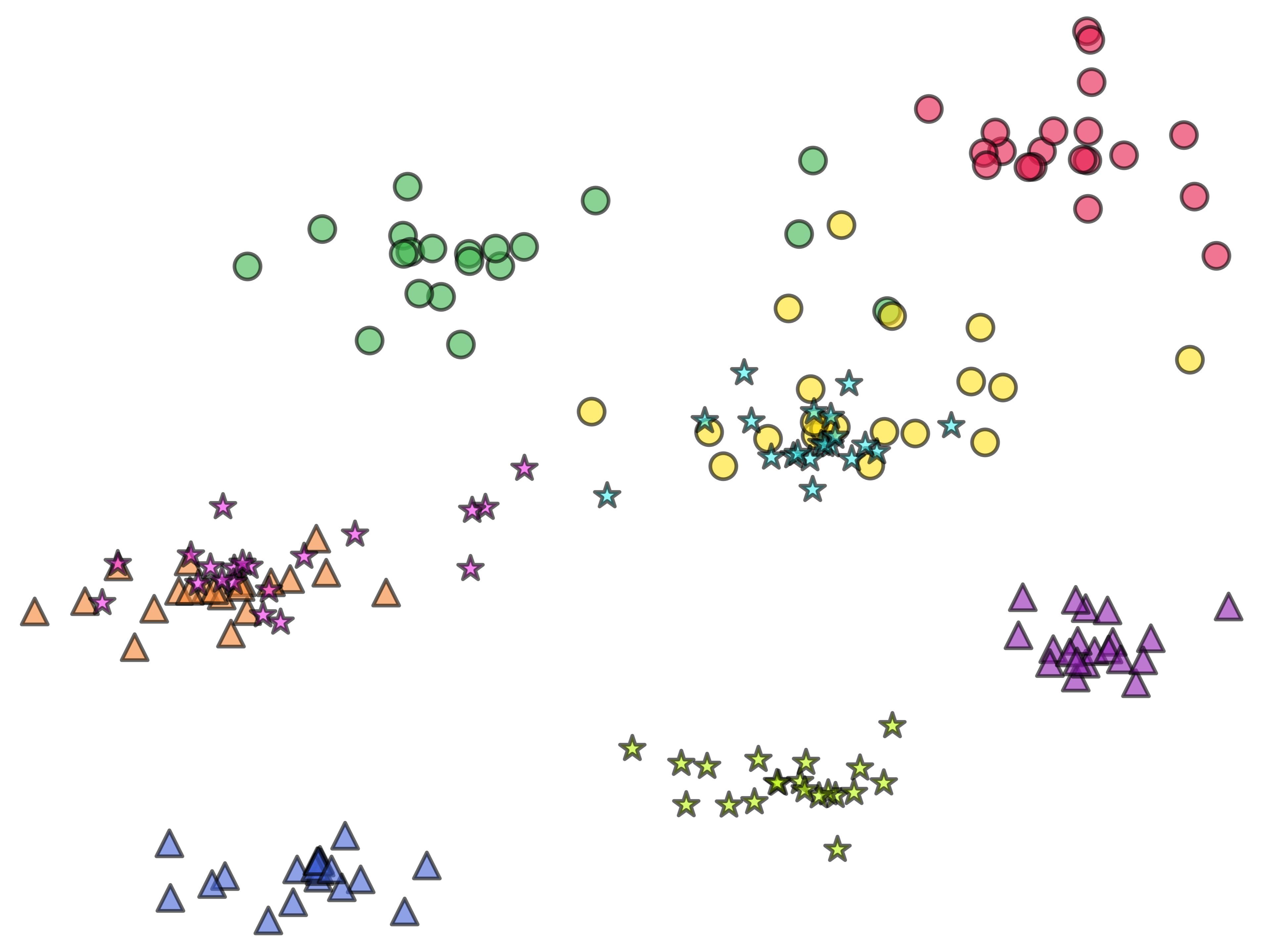}
    \caption{MQMK-EI.}
  \end{subfigure}
  \begin{subfigure}{0.48\linewidth}
    \includegraphics[width=\textwidth, trim=0 0 0 0 , clip]{figures/mq_tsne.png}
    \caption{Selected Queries in MQ.}
  \end{subfigure}
  
  \caption{t-SNE visualization of the Queries. The samples come from three classes of three tasks in the 5-task Split Imagenet-R. The shapes represent the sample tasks, and the colors represent the sample categories.}
  \label{fig:tsne_ei}
\end{figure}
As illustrated in \cref{fig:tsne_ei}, the enhanced queries in MQMK-EI and the standard queries in MQMK both demonstrate improved separability compared to SQ. Notably, MQMK shows a slightly higher degree of query separability than MQMK-EI.
These results indicate that MQMK-EI can be regarded as an efficient approximation to MQMK.

\section{Performance of True and False Prompt in MQMK}
\begin{figure}[ht]
    \centering
    \includegraphics[width=0.35\textwidth,trim=0 10 0 10,clip]{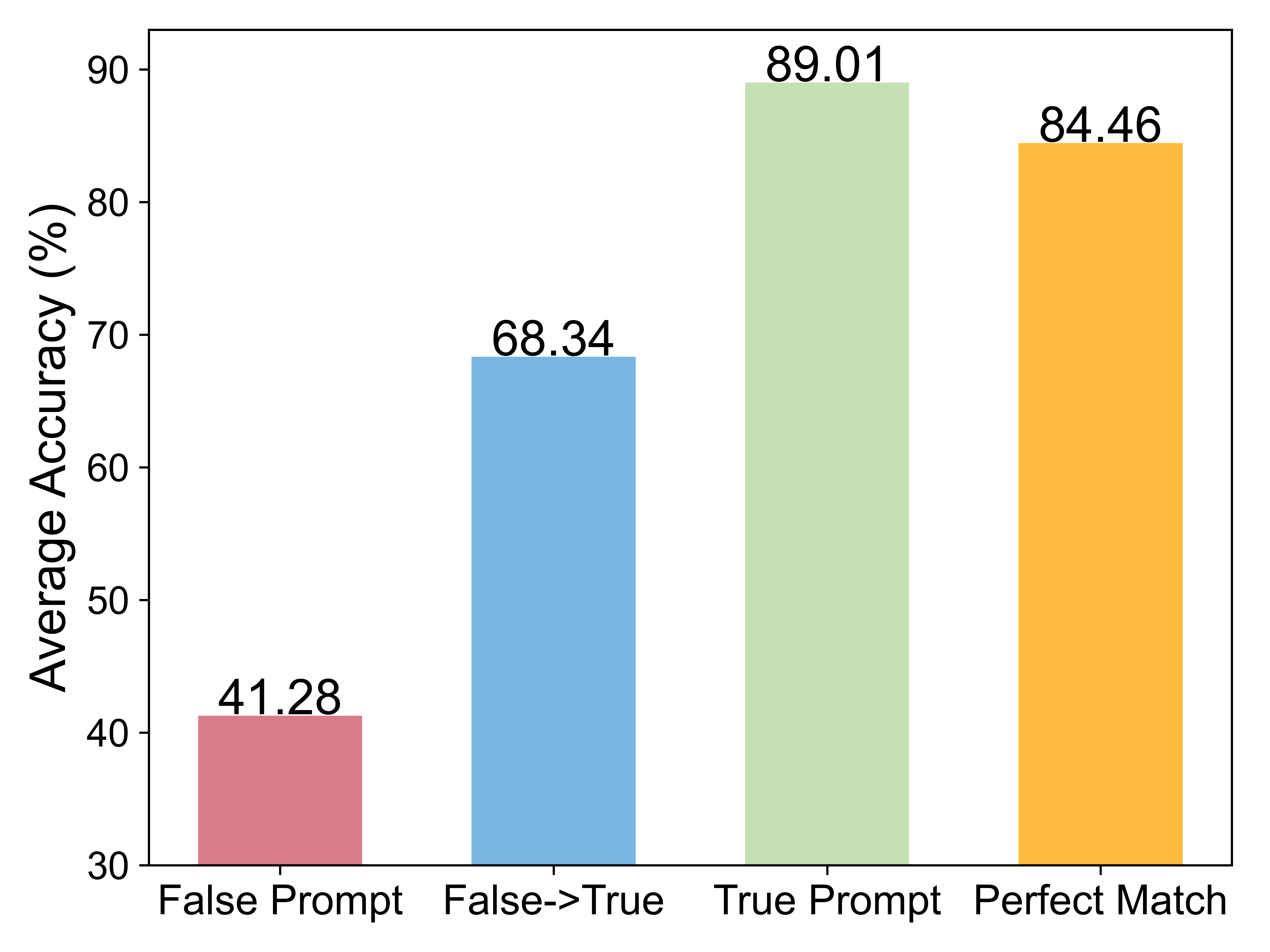}
    \caption{The average accuracy for four scenarios: when MQMK selects False Prompt and True Prompt, when samples initially with False Prompt are manually replaced with True Prompt, and when all samples use True Prompt.
    }
    \label{fig:acc_true_false_perfect_mqmk}
\end{figure}
We have already discussed the average accuracy of the samples selected with True and False Prompts in SQSK.
In this section, we will discuss the situation in MQMK and compare the two scenarios.
Compared to SQSK, MQMK has a lower accuracy for the incorrectly selected False Prompt samples.
Even when provided with the True Prompt, the performance remains only at 68.34\%.
\textbf{This suggests that MQMK has already matched the more recognizable and straightforward samples, and the remaining samples are inherently difficult to recognize.}
Even with the True Prompt, the performance is limited.
The samples selected with the True Prompt in MQMK are easier to match and recognize, achieving an accuracy of 89.01\%.

\section{Effect of Top-$K$ in Category Aggregation}
\begin{table}[ht]

\centering
\begin{tabular}{c||cc}
\hline
$K$ & $A_T$ (\(\uparrow\)) & Matching Rate \((\uparrow\))\\
\hline
1 & 78.82 & 77.97 \\

2 & 76.28  & 71.60  \\

4 & 72.19  & 59.98 \\

10 & 65.10 & 42.13 \\

20 & 59.50 & 27.58 \\

\hline
\end{tabular}
\caption{Average accuracy (\%) and matching rate (\%) under different $K$-category aggregation on 10-task Split Imagenet-R.}
\label{tab:number_k}
\end{table}

In previous experiments, we set $K$ to 1 at all times.
Here, we will discuss the performance of aggregating multiple classes and then selecting the prompt.
As shown in \cref{tab:number_k}, when $K$ is 1, both performance and matching rate reach their optimal values.
As $K$ increases, both performance and matching rate decline.
This suggests that the classes within the same task are not similar in features, and aggregating the matching scores for a group of classes has a negative effect.


\section{Dataset Introduction}
\label{sec:dataset}
In this section, we introduce three datasets that are used.
\setlist[itemize]{leftmargin=1.2em, labelwidth=0em}
\begin{itemize}
  \item CIFAR-100 is a popular dataset in machine learning, consisting of 100 different classes.
  It contains 60,000 32$\times$32 color images, divided into 50,000 training images and 10,000 test images.
  Each class contains 600 images, and the categories are diverse, such as animals, vehicles, and household items.
  \item ImageNet-R is a variant of the ImageNet dataset, designed to evaluate the robustness of machine learning models to transformed or corrupted images.
  It contains 100,000 images from 200 classes, similar to the original ImageNet, but with images altered by various transformations such as noise, blur, or weather effects.
  The dataset is useful for testing the generalization and resilience of models to real-world changes in data.
  \item DomainNet is a large-scale dataset designed to address domain adaptation tasks.
  It consists of 6 different domains (e.g., real, clipart, painting, and sketch) with a total of over 600,000 images across 345 categories.
  The dataset is used to evaluate how well models trained on one domain can transfer to others, making it ideal for research on domain adaptation and transfer learning.
  Since the data in DomainNet is primarily concentrated within 200 classes, we select the 200 classes with the most samples, using the same class-incremental setting and train-test split as in \cite{gao2024consistent}.
  Therefore, we compare our results with those reported in their paper on this dataset.
\end{itemize}

\section{Explanation of MQMK-EI and Its Forgetting Behavior}

The prompts learned for different tasks are approximately orthogonal. As the model size and the length of the prompts increase, both prompts and features reside in a very high-dimensional space, where random vectors naturally tend to be orthogonal. This approximate orthogonality among different prompts may serve as the foundation for the effectiveness of prompt fusion.

It is a common phenomenon that MQMK-EI exhibits lower forgetting rates but also lower accuracy compared to MQMK, as shown in the Performance Comparison section of the main text.
We found that after task $i$ is learned, MQMK achieves higher accuracy than MQMK-EI on the samples from that task. However, as more new tasks are learned, the accuracy of MQMK on task $i$ decreases more significantly, although it still remains higher than that of MQMK-EI, as shown in \cref{tab:task_2_Accuracy}. This is because the fusion strategy in MQMK-EI introduces interference from previously learned tasks to the current one, leading to performance degradation on the current task. In contrast, MQMK benefits from strong query independence, enabling it to achieve higher performance immediately after training. Nevertheless, as interference accumulates with the learning of more tasks, MQMK eventually experiences performance degradation as well.
Based on this analysis, the fewer the number of tasks, the greater the advantage of MQMK.






\section{Upper Bound}

\begin{table}[ht]
\centering
\begin{tabular}{l||ccc}
\hline
      &ImageNet-R&	CIFAR-100& DomainNet\\
\hline
Perfect Match &84.46	&94.91 &	91.79\\

Joint Train &79.27	& 91.79 &89.15\\

\hline
\end{tabular}
\caption{The comparison between Joint Train and Perfect Match.}
\label{tab:Matching_Rates}
\end{table}

The comparison between Joint Train and Perfect Match is shown in \cref{tab:Matching_Rates}. The accuracy of Perfect Match is significantly higher than that of Joint Training. This suggests that the matching approach still has room for improvement and holds considerable promise.



\end{document}